%% file: main.tex
\documentclass[letterpaper, 10pt, conference]{ieeeconf}
\IEEEoverridecommandlockouts   % needed for \thanks in the author block
\usepackage{cite}
\usepackage{amsmath,amssymb,amsfonts}
\usepackage{graphicx}
\usepackage{tikz}
\usetikzlibrary{positioning,arrows.meta}
\usepackage{booktabs}
\usepackage{algorithm}
\usepackage{algpseudocode}
\usepackage{multirow}
\usepackage{array}
\usepackage{xcolor}
\usepackage{listings}
\usepackage{url}
\usepackage[hidelinks]{hyperref}  % loaded last

\providecommand{\IEEEPARstart}[2]{#1#2}  % drop the journal drop-cap
\makeatletter
\@ifundefined{IEEEkeywords}{%
  \newsavebox{\discard@keywords}%
  \newenvironment{IEEEkeywords}%
    {\begin{lrbox}{\discard@keywords}\begin{minipage}{\linewidth}}%
    {\end{minipage}\end{lrbox}}%
}{}
\makeatother

\definecolor{codekw}{HTML}{0B5394}
\definecolor{codecomment}{HTML}{6B7280}
\definecolor{codestring}{HTML}{15803D}
\definecolor{codeemph}{HTML}{B45309}
\definecolor{codepunct}{HTML}{9AA1AC}
\lstdefinestyle{pluginapi}{
  language=Python,
  basicstyle=\scriptsize\ttfamily,
  keywordstyle=\color{codekw}\bfseries,
  commentstyle=\color{codecomment}\itshape,
  stringstyle=\color{codestring},
  emph={Backend,MyVLA},
  emphstyle=\color{codeemph},
  showstringspaces=false,
  columns=fullflexible,
  keepspaces=true,
}
\lstdefinestyle{yamlcfg}{
  basicstyle=\scriptsize\ttfamily,
  commentstyle=\color{codecomment}\itshape,
  showstringspaces=false,
  columns=fullflexible,
  keepspaces=true,
  comment=[l]{\#},
  escapeinside={(*}{*)},
  aboveskip=0pt,
  belowskip=0pt,
}
\newcommand{\ck}[1]{\textcolor{codekw}{#1}}
\newcommand{\ckv}[1]{\textcolor{codestring}{#1}}
\newcommand{\ckn}[1]{\textcolor{codeemph}{#1}}
\newcommand{\ckp}[1]{\textcolor{codepunct}{#1}}
\newsavebox{\apibox}
\newlength{\apih}

\newcommand{\pifive}{$\pi_{0.5}$}
\newcommand{\decentvla}{\textsc{decent-vla}}

\begin{document}

\title{\LARGE \bf An Empirical Study and Open Testbed for Federated
Fine-Tuning of Vision-Language-Action Models}

\author{Zhekai Duan$^{1,*}$, Kevin Ziyang Xie$^{1}$, Xinyu Tan$^{1}$,
Shikai Geng$^{1}$, Chengxu Zhou$^{1}$,\\
Ramana Kompella$^{2}$, Gaowen Liu$^{2}$, and Chris Xiaoxuan Lu$^{1}$%
\thanks{$^{1}$Department of Computer Science, University College London, U.K.}%
\thanks{$^{2}$Cisco Research, San Jose, CA, USA.}%
\thanks{$^{*}$Corresponding author: \texttt{zhekaiduan2312@gmail.com}}%
}

\maketitle
\thispagestyle{empty}
\pagestyle{empty}

\input{sections/abstract}          % keywords are discarded by the shim above

\input{sections/introduction}
\input{sections/related_work}
\input{sections/problem_formulation}
\input{sections/framework}
\input{sections/benchmark_design}
\input{sections/experiments}
\input{sections/conclusion}
\input{sections/acknowledgment}
\input{sections/references}

\end{document}

%% file: sections/abstract.tex
\begin{abstract}
Adapting a pretrained Vision-Language-Action (VLA) model to a new robot,
environment, or task requires demonstrations that are collected locally
and often discarded. Federated learning is a promising approach to exploiting such 
distributed demonstrations by learning a shared policy. However, whether it can adapt large pretrained VLAs remains
an open question, and a lack of reproducible benchmarks for pretrained
VLAs and reusable training frameworks makes existing results difficult
to compare. In this paper, we conduct a systematic study of federated fine-tuning of
three modern pretrained VLA policies on the 40 simulated tasks of the
LIBERO manipulation benchmark, and on six real-world tasks in two real-robot experiments, with demonstrations
collected across two and three sites, respectively. Our study analyzes the key choices in this
setting, spanning multiple federated parameter scopes, three aggregation
algorithms, and evaluation under distribution
shift. Based on the study, we derive a series of lessons, including the
dominance of the federated scope over the choice of aggregation
algorithm and the difficulty of matching centralized fine-tuning on
physical robots, where cross-site heterogeneity is stronger than
simulation captures. We also highlight opportunities for
federated VLA learning, such as the ability to match centralized
fine-tuning on heterogeneous data, to remain at least as robust as
centralized fine-tuning under
distribution shift, and to personalize, with each client federating part
of the policy and keeping the rest local, which helps where the
policy's pretraining is weak but leaves no usable global model. We open-source \decentvla{},
the model- and runtime-agnostic testbed behind the study, to facilitate
future research and fair comparisons in federated VLA learning.
% Held out for ICRA 2027 double-anonymous review (the username identifies
% the authors); restore for the camera-ready version.
% Code and the configuration behind every reported number are available at
% \url{https://github.com/kevinDuan1/decent-vla}.
\end{abstract}


%% file: sections/introduction.tex
\section{Introduction}\label{sec:intro}
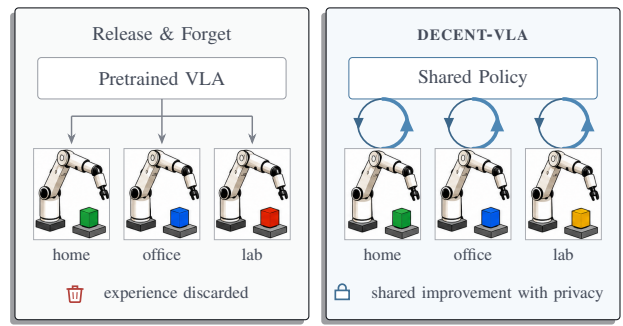
\begin{figure}[t]
\centering
% Normalized to the column measure so the fixed-cm drawing fits both the
% IEEEtran (RA-L) and ieeeconf (ICRA) column widths.
\resizebox{\columnwidth}{!}{\input{figures/openfig.tikz}}
\caption{In current practice (left), each site downloads a released VLA checkpoint, and the demonstrations it collects afterwards go unused. \decentvla{} (right) lets each site fine-tune the shared
policy on its private data and exchange only model updates, so the
policy keeps improving while data stays local.}
\label{fig:motivation}
\end{figure}

\IEEEPARstart{V}{ision}-Language-Action (VLA) models map visual observations
and natural-language instructions to robot actions, and openly released
checkpoints such as \pifive{}~\cite{pi05}, SmolVLA~\cite{smolvla}, and
GR00T~N1.7~\cite{grootn17} (hereafter GR00T) have made pretrained
generalist policies broadly available. Unlike the language and vision
models they build on, a pretrained VLA must still be fine-tuned for
each new robot, environment, or task, on demonstrations collected at
the deployment site. Today this adaptation runs in one direction:
each site fine-tunes the released checkpoint in isolation, and the
demonstrations collected during deployment are discarded
(Fig.~\ref{fig:motivation}). What is missing is the reverse
mechanism: a way for many holders of small, private, heterogeneous
datasets to \emph{collectively} improve a shared open policy.

Federated learning (FL)~\cite{fedavg} provides this mechanism: each site
fine-tunes the policy locally and transmits only model updates to an
aggregating server, an approach recently explored for large language
models (LLMs)~\cite{fedit,openfedllm}. VLA data, however, is heterogeneous
along axes with no analogue in text or images. Even two sites running the
same robot model can differ not only in their tasks, but also in their
observations, shaped by environment and camera viewpoint, and in their
actions and states, shaped by hardware calibration and operator
behavior. Whether federated
optimization can adapt a large pretrained VLA across such sites while
preserving the capabilities acquired during pretraining is an open question.

Existing studies~\cite{fedvla, flame} use different policies, partitions, aggregation algorithms, and evaluation setups,
so their results are difficult to compare, and each rebuilds its
training infrastructure from scratch. The field lacks a reusable
training framework and a reproducible benchmark on which a new
policy, aggregation algorithm, or runtime can be evaluated under
identical conditions.

We therefore conduct a systematic study and build the testbed it requires.
\textbf{\decentvla{}} is an open \emph{testbed} with two parts: a
\emph{framework} that federates the fine-tuning of pretrained VLAs,
in which an experiment is specified by a configuration file and a
new policy, aggregation algorithm, or runtime is added as a small
plugin (Sec.~\ref{sec:framework}); and a \emph{reference benchmark}
that fixes the partition, training budgets, and evaluation protocol
(Sec.~\ref{sec:benchmark}). On this testbed we study three
modern pretrained policies on LIBERO in simulation, and \pifive{} on
real SO-101 robot arms at two and three sites.

\smallskip
\noindent\textbf{Contributions.}
\begin{enumerate}
  \item \textbf{Study.} A systematic empirical study of federated
        fine-tuning of \pifive{}, SmolVLA, and GR00T~N1.7, on LIBERO
        in simulation and on real SO-101 arms. Federated fine-tuning
        can match centralized fine-tuning when the policy's
        pretraining is strong; the federated scope, not the
        aggregation algorithm, governs the outcome; and the
        calibration heterogeneity of real sites keeps centralized
        fine-tuning ahead (Sec.~\ref{sec:experiments}).
  \item \textbf{Testbed.} \decentvla{}, an easy-to-use testbed that
        supports different policies, aggregation algorithms, and FL
        runtimes; the configuration behind every reported number in
        the study is released (Sec.~\ref{sec:framework}).
\end{enumerate}

%% file: figures/openfig.tikz
\definecolor{shadeink}{HTML}{16191F}
\definecolor{stageink}{HTML}{444B59}
\definecolor{flcol}{HTML}{5C9FD4}     % federated -- as in real.tikz/purt.tikz
\definecolor{binred}{HTML}{B3413A}    % nocol, the family's negative-status red

\begin{tikzpicture}[x=1cm, y=1cm,
  soft/.style={fill=shadeink, fill opacity=0.42, draw=none, rounded corners=1.5pt},
  mountL/.style={fill=stageink!3, draw=shadeink, line width=0.45pt,
                 rounded corners=1.5pt},
  mountR/.style={fill=flcol!6,   draw=shadeink, line width=0.45pt,
                 rounded corners=1.5pt},
  frame/.style={draw=stageink!45, line width=0.3pt},
  chipL/.style={fill=white, draw=stageink!55, line width=0.4pt,
                rounded corners=1.5pt},
  chipR/.style={fill=white, draw=flcol!75!black, line width=0.4pt,
                rounded corners=1.5pt},
  ttl/.style={font=\footnotesize, text=stageink, anchor=base},
  box/.style={font=\footnotesize, text=stageink, anchor=center},
  sml/.style={font=\scriptsize, text=stageink, anchor=base},
  foot/.style={font=\scriptsize, text=stageink, anchor=base},
  flowL/.style={stageink!70, line width=0.6pt},
  flowR/.style={flcol!70!black, line width=0.6pt},
  hd/.style={-{Stealth[length=1.6mm, width=1.2mm]}},
  % The two halves are plain strokes; their heads are drawn by hand below.
  % Weights are the argument of the paper: the broadcast is the thin half (it is
  % what already happens today, and the left panel has it too), the RETURN is the
  % heavy half. Do not equalise them.
  down/.style={flcol!65!black, line width=0.7pt},
  back/.style={flcol!85!black, line width=1.5pt, line cap=round},
  headd/.style={fill=flcol!65!black, draw=none},
  headb/.style={fill=flcol!85!black, draw=none},
  % --- icons, verbatim from motivation.tikz -------------------------------
  pics/lock/.style={code={%
    \draw[#1, line width=0.7pt, fill=white] (-0.10,-0.075) rectangle (0.10,0.055);
    \draw[#1, line width=0.7pt]
      (-0.055,0.055) -- (-0.055,0.09)
      arc[start angle=180, end angle=0, radius=0.055] -- (0.055,0.055);
  }},
  pics/bin/.style={code={%
    \draw[#1, line width=0.7pt, line cap=round] (-0.115,0.075) -- (0.115,0.075);
    \draw[#1, line width=0.7pt]
      (-0.035,0.075) -- (-0.035,0.115) -- (0.035,0.115) -- (0.035,0.075);
    \draw[#1, line width=0.7pt, fill=white, rounded corners=0.4pt]
      (-0.095,0.075) -- (-0.075,-0.115) -- (0.075,-0.115)
      -- (0.095,0.075) -- cycle;
    \draw[#1, line width=0.45pt] (-0.030,0.030) -- (-0.025,-0.070);
    \draw[#1, line width=0.45pt] (0.030,0.030) -- (0.025,-0.070);
  }},
]

% Panel-local coordinates are used throughout: each panel is drawn inside a
% scope shifted to its own origin, so the two layouts are literally the same
% numbers and can never drift apart.
\def\PW{4.26}     % panel width
\def\PH{4.52}     % panel height
\def\IW{1.10}     % image tile width
\def\IH{1.321}    % ... and height, the crops' 194:233 aspect
\def\TY{1.16}     % tile baseline (bottom edge)

% ===== left panel: release and forget =======================================
\begin{scope}[shift={(0,0)}]
  \fill[soft]  (-0.07,-0.07) rectangle ({\PW-0.07},{\PH-0.07});
  \draw[mountL] (0,0) rectangle (\PW,\PH);

  \node[ttl] at ({\PW/2},4.06) {Release \& Forget};
  \draw[chipL] (0.33,3.20) rectangle (3.93,3.80);
  \node[box]  at ({\PW/2},3.50) {Pretrained VLA};

  % one-way fan-out: down to a rail, then a head into each site
  \draw[flowL] ({\PW/2},3.20) -- ({\PW/2},2.95);
  \draw[flowL] (0.82,2.95) -- (3.44,2.95);
  \foreach \cx in {0.82,2.13,3.44}
    \draw[flowL, hd] (\cx,2.95) -- (\cx,{\TY+\IH+0.06});

  \foreach \cx/\img in {0.82/green-l, 2.13/blue-l, 3.44/red-l} {
    \node[anchor=south west, inner sep=0] at ({\cx-\IW/2},\TY)
      {\includegraphics[width=\IW cm, height=\IH cm]
        {figures/photos/openfig/arm-\img.png}};
    \draw[frame] ({\cx-\IW/2},\TY) rectangle ({\cx+\IW/2},{\TY+\IH});
  }
  \foreach \cx/\s in {0.82/home, 2.13/office, 3.44/lab}
    \node[sml] at (\cx,{\TY-0.30}) {\s};

  \node[foot] (fl) at ({\PW/2+0.21},0.31) {experience discarded};
  \pic at ([xshift=-0.30cm, yshift=0.06cm]fl.base west) {bin=binred};
\end{scope}

% ===== right panel: decent-vla closes the loop ==============================
\begin{scope}[shift={(4.51,0)}]
  \fill[soft]  (-0.07,-0.07) rectangle ({\PW-0.07},{\PH-0.07});
  \draw[mountR] (0,0) rectangle (\PW,\PH);

  \node[ttl] at ({\PW/2},4.06) {\textbf{\decentvla{}}};
  \draw[chipR] (0.33,3.20) rectangle (3.93,3.80);
  \node[box]  at ({\PW/2},3.50) {Shared Policy};

  % One wheel per site rather than a pair of straight rails: the broadcast and
  % the return are the two halves of the SAME CIRCLE, so each site reads as a
  % closed round rather than as two unrelated arrows. Down the left side, back
  % up the right, the direction a reader already scans a cycle.
  %
  % A true circle, both radii 0.36, not an ellipse: r is set by the gap between
  % the chip (3.20) and the tile tops (2.48), which is 0.72 = 2r exactly, so the
  % wheels touch both without being squeezed. Circle half-width 0.36 sits inside
  % the 0.55 tile half-width, so no wheel reaches its neighbour. Changing the
  % chip or tile y means recomputing r; keep the two radii equal when you do.
  %
  % The weights are the argument of the paper. The broadcast is the thin half --
  % it is what already happens today, and the left panel has it too. The RETURN
  % is the heavy half with the large head: it is the mechanism this work adds,
  % and the sentence the introduction builds to is that it is what is missing.
  % Do not equalise these two weights.
  % Heads are drawn as explicit triangles rather than as arrow tips carried by
  % decorations.markings. \arrow puts the TIP at the marked position and hangs
  % the body off the outside of a curve, so on r=0.36 the head visibly stopped
  % sitting on the line. Built from the circle's own geometry instead, each head
  % straddles the rim: centre C=(cx,2.84), and at the 3 and 9 o'clock marks the
  % tangent is vertical, so the triangle is tip +-L/2 along y and its back
  % corners +-W either side of the rim. Centred by construction, at any radius.
  \foreach \cx in {0.82,2.13,3.44} {
    \draw[down] (\cx,3.20) arc[start angle=90,  end angle=270, radius=0.36];
    \draw[back] (\cx,2.48) arc[start angle=-90, end angle=90,  radius=0.36];
    % Each head is nudged INWARD by half its sagitta -- a straight triangle is a
    % chord across the rim, so centring its bounding box on r=0.36 leaves the
    % tip and the back corners sitting proud of the curve. Half the sagitta puts
    % the chord's own midline on the rim, which is what reads as centred.
    % broadcast, 9 o'clock, travelling down  (L 0.17, W 0.075, sagitta 0.010)
    \fill[headd] ({\cx-0.350},2.755) -- ({\cx-0.425},2.925)
              -- ({\cx-0.275},2.925) -- cycle;
    % return, 3 o'clock, travelling up      (L 0.22, W 0.100, sagitta 0.017)
    \fill[headb] ({\cx+0.351},2.950) -- ({\cx+0.251},2.730)
              -- ({\cx+0.451},2.730) -- cycle;
  }

  \foreach \cx/\img in {0.82/green-r, 2.13/blue-r, 3.44/yellow-r} {
    \node[anchor=south west, inner sep=0] at ({\cx-\IW/2},\TY)
      {\includegraphics[width=\IW cm, height=\IH cm]
        {figures/photos/openfig/arm-\img.png}};
    \draw[frame] ({\cx-\IW/2},\TY) rectangle ({\cx+\IW/2},{\TY+\IH});
  }
  \foreach \cx/\s in {0.82/home, 2.13/office, 3.44/lab}
    \node[sml] at (\cx,{\TY-0.30}) {\s};

  \node[foot] (fr) at ({\PW/2+0.21},0.31) {shared improvement with privacy};
  \pic at ([xshift=-0.30cm, yshift=0.06cm]fr.base west) {lock=flcol!70!black};
\end{scope}

\end{tikzpicture}

%% file: sections/related_work.tex
\section{Related Work}\label{sec:related}
% FLAME venue still to confirm (cited as preprint); see references.tex.

\subsection{Vision-Language-Action Models}
Modern Vision-Language-Action models adapt pretrained Vision-Language Models (VLMs) into
robot controllers by mapping visual observations and language instructions to
low-level actions. RT-2 casts actions as language tokens~\cite{rt2}, while
OpenVLA couples a 7\,B Llama-2 backbone with DINOv2 and SigLIP encoders, trained
on roughly 970\,k Open X-Embodiment demonstrations~\cite{openvla,oxe}. A second
family separates perception from control: $\pi_0$ attaches a flow-matching
action head (its ``action expert'') to a VLM~\cite{pi0}, and $\pi_{0.5}$ adds open-world generalization through
heterogeneous co-training~\cite{pi05}. SmolVLA targets single-GPU budgets
with a 0.45\,B SmolVLM2 backbone~\cite{smolvla}, and GR00T~N1 pairs a vision
language System~2 with a diffusion-transformer (DiT) action System~1~\cite{groot}. We
benchmark $\pi_{0.5}$, SmolVLA, and GR00T~N1.7, whose Cosmos-Reason2
backbone replaces the Eagle-2 of N1~\cite{grootn17}, all pretrained by centralizing large-scale, video-heavy
datasets~\cite{oxe,droid}. \decentvla{} treats each such policy as an
interchangeable plugin and instead studies how their adaptation data
can be exploited without centralization.

\subsection{Federated Learning}
In \emph{federated learning}, clients exchange model updates rather than
data: FedAvg established weighted parameter
averaging~\cite{fedavg}; heterogeneity-robust variants add a
proximal term~\cite{fedprox} or control variates~\cite{scaffold}, and
adaptive server optimizers treat aggregated updates as
pseudo-gradients~\cite{fedopt}. We benchmark FedAvg, FedProx, and
FedAdam under one benchmark configuration (Sec.~\ref{sec:q-rounds}), and \decentvla{}
drives the production systems Flower~\cite{flower} and
NVFlare~\cite{nvflare} as interchangeable execution backends rather than
competing with them (Sec.~\ref{sec:framework}). The closest line of
work federates the fine-tuning of large pretrained models through
compact updates (FedIT~\cite{fedit}, OpenFedLLM~\cite{openfedllm},
FedSA-LoRA~\cite{fedsalora}), but it federates a single modality,
language, while the federated systems that do cross modalities train
small policies or classifiers from scratch. \decentvla{} carries this
line from language to the adaptation of large pretrained VLAs, over a
federated scope that can cover either the full model or a
parameter-efficient adapter such as low-rank adaptation
(LoRA)~\cite{lora}. A complementary line personalizes the federated
model by keeping part of it local to each client, through
personalization layers~\cite{fedper}, local batch
normalization~\cite{fedbn}, or bi-level objectives~\cite{ditto}. We
study the same split at VLA scale, federating one scope of the policy
while each client fine-tunes the remainder locally.

On the robotics side, fleet-learning systems aggregate experience
across many robots~\cite{lfrl,fleetmerge,openbotfleet}, but do not
fine-tune pretrained VLAs. The two closest efforts each miss the
intersection we target, a systematic study across multiple modern
pretrained VLAs: FLAME federates a small from-scratch policy in
simulation only~\cite{flame}, and FedVLA is a single method on one
bespoke backbone rather than a benchmark~\cite{fedvla}. Spanning
multiple pretrained policies, aggregation algorithms, and both
a heterogeneous LIBERO partition and a multi-site study on real SO-101
arms, \decentvla{} reveals effects that neither line of work measured:
aggregation algorithms designed for heterogeneous clients that do not
improve on FedAvg at a matched round budget (Sec.~\ref{sec:q-rounds}), and a calibration
heterogeneity that appears only when the clients are physical robots
(Sec.~\ref{sec:q-real}).

% --- framework-comparison table ------------------------------------------
% \begin{table}[t]
% \caption{Positioning of \decentvla{} against representative federated-learning
% systems and the two closest federated-robotics efforts.}
% \label{tab:frameworks}
% \centering
% \footnotesize
% \renewcommand{\arraystretch}{1.25}
% \setlength{\tabcolsep}{4pt}
% \begin{tabular}{lccccc}
% \toprule
%  & Pre-tr. & Multi- & Runtime- & Sim\,\& & Rel. \\
%  & VLA & policy & agnostic & Real & bench. \\
% \midrule
% Flower~\cite{flower}         & no  & ---     & ---  & ---  & no \\
% NVFlare~\cite{nvflare}       & no  & ---     & ---  & ---  & no \\
% OpenFedLLM~\cite{openfedllm} & LLM & partial & no   & ---  & part. \\
% FLAME~\cite{flame}           & no  & no      & no   & sim  & yes \\
% FedVLA~\cite{fedvla}         & yes & no      & no   & both & no \\
% \textbf{\decentvla{}}        & \textbf{yes} & \textbf{yes} & \textbf{yes} & \textbf{both} & \textbf{yes} \\
% \bottomrule
% \end{tabular}
% \end{table}

%% file: sections/problem_formulation.tex
\section{Problem Setup: Variables and Questions}\label{sec:problem}
We consider $K$ sites (clients) that each hold a private
demonstration dataset
$\mathcal{D}_k$ of $n_k$ samples that cannot leave the site. All
clients start from a common pretrained VLA $\theta^0$, and federated
fine-tuning seeks a single global policy that performs well on the
union of their tasks. Algorithm~\ref{alg:protocol} summarizes the
process of federated fine-tuning: in each of
$R$ rounds, the server broadcasts the parameters in the federated
scope $\mathcal{S}$, each client updates them on its local data, and
the server aggregates the returned updates into the global
parameters. We outline five questions about federated fine-tuning
of pretrained VLAs:

\begin{algorithm}[t]
\caption{Federated fine-tuning of a pretrained VLA.}
\label{alg:protocol}
\begin{algorithmic}[1]
\Require pretrained $\theta^{0}$; clients $k=1,\dots,K$ with private
  $\mathcal{D}_k$ ($n_k$ samples); scope $\mathcal{S}$; rounds $R$
\For{$r = 1,\dots,R$}
  \State server broadcasts $\theta^{r-1}[\mathcal{S}]$ to all clients
  \For{client $k = 1,\dots,K$ \textbf{in parallel}}
    \State $\theta_k \gets$ local imitation steps on $\mathcal{D}_k$
      starting from $\theta^{r-1}$
    \State upload $\theta_k[\mathcal{S}]$ to the server
  \EndFor
  \State $\theta^{r}[\mathcal{S}] \gets
    \textsc{Aggregate}\bigl(\{\theta_k[\mathcal{S}]\}_{k=1}^{K}\bigr)$
\EndFor
\end{algorithmic}
\end{algorithm}

% --- wide architecture figure (belongs to Sec. IV-A) ----------------------
% Sourced here, one page ahead of its first reference, because a
% double-column float can only be placed on a page after the one on
% which it is encountered; this puts it on the page where Sec. IV starts.
\begin{figure*}[t]
\centering
\input{figures/arch.tikz}
\caption{The \decentvla{} architecture. Each round, the server
broadcasts the model parameters $\theta$ (restricted to the federated
scope), each client trains them on its local data and returns its
update $\Delta\theta$ with its sample count $n$, and a pluggable
\emph{Strategy} aggregates the updates. Every client composes a
\emph{policy backend} (the VLA) and an \emph{execution backend} (the
FL runtime) independently, so the same loop runs in-process or over
Flower~\cite{flower} or NVFlare~\cite{nvflare}.}
\label{fig:arch}
\end{figure*}
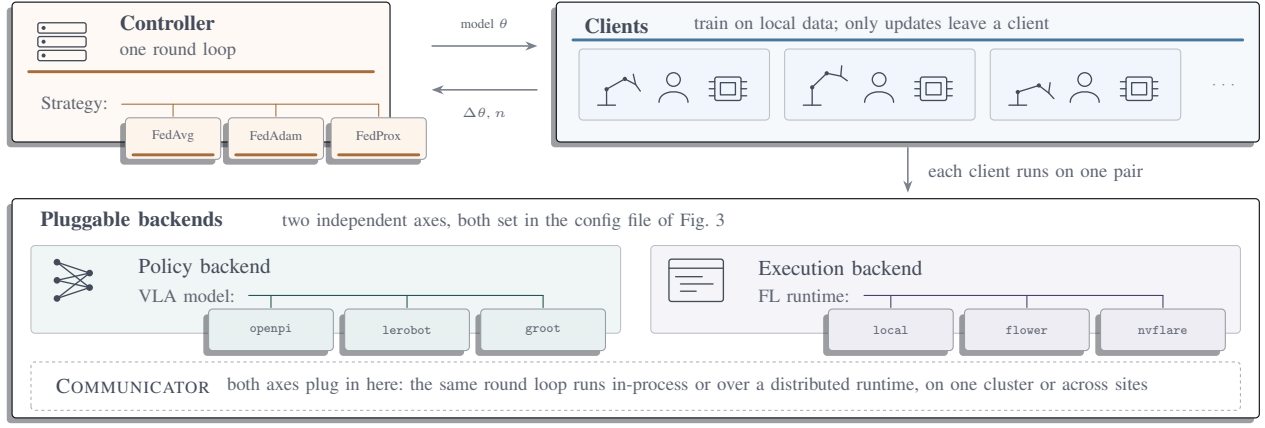

\paragraph*{Federated versus centralized fine-tuning (Q1)}
Robot demonstration data is strongly heterogeneous: sites differ in
their tasks, environments, observations, and actions. Federated
optimization is known to trail centralized training under such
heterogeneity, as local updates drift toward each client's
objective~\cite{fedavg,scaffold}, but the size of the gap for
billion-parameter pretrained VLAs has not been measured. We assign
each client a disjoint subset of tasks and measure how closely federated
fine-tuning approaches centralized fine-tuning
(Sec.~\ref{sec:q-gap}).

\paragraph*{Federated scope $\mathcal{S}$ (Q2)}
VLAs can have billions of parameters, so exchanging the full model
between server and clients every round, ${\sim}7$\,GB in
\texttt{bf16} for \pifive{}, limits the practicality of
federated fine-tuning. Federating only part of the model reduces
this payload, and it also enables personalization: the parameters
outside $\mathcal{S}$ can be fine-tuned locally and kept personal to
each client (\emph{partial} federation)~\cite{fedper,ditto}. We
therefore investigate LoRA~\cite{lora} updates in place of
full fine-tuning and the module-wise federation of the action head
or the backbone (Sec.~\ref{sec:q-scope}).

\paragraph*{Aggregation algorithm (Q3)}
Alternatives to FedAvg~\cite{fedavg} designed for heterogeneous
clients, such as the proximal term of FedProx~\cite{fedprox} and the
server-side adaptivity of FedAdam~\cite{fedopt}, are reported to
improve performance under precisely the heterogeneity that robot
data presents, but remain untested on large pretrained VLAs. We
compare the performance of the three algorithms
(Sec.~\ref{sec:q-rounds}).

\paragraph*{Robustness under distribution shift (Q4)}
Pretrained VLAs degrade substantially when
instructions, objects, scene layouts, or goals shift at evaluation
time~\cite{liberopro}, and whether federated fine-tuning is more or
less robust to such shifts than centralized fine-tuning is unknown. We
evaluate the federated and centralized fine-tuned VLAs under four axes of
evaluation-time shift (Sec.~\ref{sec:q-robust}).

\paragraph*{Transfer to physical robots (Q5)}
On physical robots, client heterogeneity extends beyond the task differences
that simulated benchmarks express: sites differ in their observations,
shaped by environment and camera viewpoint, and in their actions and
states, shaped by hardware calibration and operator behavior. This
heterogeneity has no analogue in federated language
models~\cite{fedit,openfedllm}. We compare federated and centralized fine-tuning on
real SO-101 arms at two and three sites, then rescore the same
checkpoints in a simulated environment calibrated to one site for
repeatable layouts and stage-wise success rates (Sec.~\ref{sec:q-real}).

%% file: figures/arch.tikz
\definecolor{shadeink}{HTML}{16191F}
\definecolor{stageink}{HTML}{444B59}
\definecolor{flcol}{HTML}{5C9FD4}    % federated   -- must track stage.tikz
\definecolor{centcol}{HTML}{EC9A55}  % centralized -- must track stage.tikz
% The two pluggable axes get hues of their OWN, not the reserved pair: policy
% and execution are not federated and centralized, and borrowing blue/orange
% here would collide with the results figures (see the monochrome note above,
% which this respects rather than overturns -- the axes are still not drawn in
% the series colours). Teal and violet are far from both #5C9FD4 and #EC9A55
% and from each other, and are used only as a panel tint, a chip fill and the
% leader that ties a label to its options.
\definecolor{polcol}{HTML}{2F7D7A}   % policy axis    -- teal
\definecolor{execol}{HTML}{6E5B9E}   % execution axis -- violet

\begin{tikzpicture}[x=1cm, y=1cm,
  soft/.style={fill=shadeink, fill opacity=0.42, draw=none, rounded corners=1.5pt},
  panel/.style={fill=white, draw=shadeink, line width=0.45pt, rounded corners=1.5pt},
  % Tinted panels. The monochrome rule below still holds for the two pluggable
  % axes -- policy and execution are not federated and centralized, so they
  % stay neutral. But the top row IS that distinction: the controller is the
  % single central point and the clients are the federated side, and the
  % accent rules already said so in line. These carry the same meaning into
  % the panel fill so the figure reads in colour at a glance.
  panelsrv/.style={panel, fill=centcol!6},
  panelfed/.style={panel, fill=flcol!6},
  sub/.style={fill=stageink!5, draw=stageink!35, line width=0.4pt, rounded corners=1.5pt},
  site/.style={draw=stageink!45, line width=0.4pt, dashed, dash pattern=on 1.2pt off 1.2pt,
               rounded corners=1.5pt},
  chip/.style={fill=white, draw=stageink!55, line width=0.4pt, rounded corners=1.5pt},
  ic/.style={stageink, line width=0.5pt, line cap=round, line join=round},
  icf/.style={stageink, fill=stageink},
  ttl/.style={font=\footnotesize, text=stageink, anchor=west},
  sml/.style={font=\scriptsize, text=stageink!85, anchor=west},
  tny/.style={font=\tiny, text=stageink!85, anchor=center},
  flow/.style={stageink!70, line width=0.6pt, -{Stealth[length=1.6mm, width=1.2mm]}},
  % Leader from a label to the row of options it names: a spine at the
  % label's own baseline with a short drop onto each chip. The label cannot
  % sit in line with the chips -- see the note by `Strategy:' -- so the
  % spine is what ties the two together.
  lead/.style={draw=stageink!55, line width=0.4pt, line cap=round},
  % Accents, not encodings: they mark which half of the picture is the
  % federated machinery and which is the single central point. Darker steps
  % of the series hues, because the base pair sits near 2.2--2.8:1 against
  % white and a 1pt rule at that lightness disappears in print.
  accfed/.style={flcol!75!black, line width=1.1pt, line cap=round},
  accsrv/.style={centcol!70!black, line width=1.1pt, line cap=round},
]

% ===== icon macros ==========================================================
% Each draws inside a 1x0.8 box at the current origin; wrap in a shifted,
% scaled scope to place one.
\newcommand{\icserver}{%
  \foreach \y in {0.06,0.30,0.54}{
    \draw[ic, rounded corners=0.8pt] (0.08,\y) rectangle (0.92,{\y+0.18});
    \fill[icf] (0.18,{\y+0.09}) circle (0.035);}}
\newcommand{\icnet}{%
  \foreach \y in {0.10,0.40,0.70}{\foreach \z in {0.28,0.52}{\draw[ic] (0.18,\y) -- (0.72,\z);}}
  \foreach \y in {0.10,0.40,0.70}{\fill[icf] (0.18,\y) circle (0.055);}
  \foreach \z in {0.28,0.52}{\fill[icf] (0.72,\z) circle (0.055);}}
\newcommand{\icscreen}{%
  \draw[ic, rounded corners=1pt] (0.05,0.05) rectangle (0.95,0.75);
  \draw[ic] (0.05,0.60) -- (0.95,0.60);
  \draw[ic] (0.16,0.46) -- (0.60,0.46);
  \draw[ic] (0.24,0.34) -- (0.78,0.34);
  \draw[ic] (0.24,0.22) -- (0.50,0.22);}
% Three arm poses, not one repeated. The paper's heterogeneity axes are
% "tasks ... hardware calibration, physical environment, and operator
% behavior" (Sec. I), so identical client cards would contradict the text.
% Pose stands for calibration; the operator glyph stands for the fourth axis.
\newcommand{\icarmA}{%
  \draw[ic] (0.14,0.06) -- (0.52,0.06);
  \draw[ic] (0.33,0.06) -- (0.33,0.40);
  \draw[ic] (0.33,0.40) -- (0.72,0.62);
  \fill[icf] (0.33,0.40) circle (0.045);
  \fill[icf] (0.72,0.62) circle (0.045);
  \draw[ic] (0.72,0.62) -- (0.95,0.50);
  \draw[ic] (0.95,0.50) -- (0.90,0.72);
  \draw[ic] (0.95,0.50) -- (1.05,0.36);}
\newcommand{\icarmB}{%
  \draw[ic] (0.14,0.06) -- (0.52,0.06);
  \draw[ic] (0.33,0.06) -- (0.33,0.44);
  \draw[ic] (0.33,0.44) -- (0.64,0.74);
  \fill[icf] (0.33,0.44) circle (0.045);
  \fill[icf] (0.64,0.74) circle (0.045);
  \draw[ic] (0.64,0.74) -- (0.94,0.68);
  \draw[ic] (0.94,0.68) -- (0.96,0.90);
  \draw[ic] (0.94,0.68) -- (1.08,0.58);}
\newcommand{\icarmC}{%
  \draw[ic] (0.14,0.06) -- (0.52,0.06);
  \draw[ic] (0.33,0.06) -- (0.33,0.34);
  \draw[ic] (0.33,0.34) -- (0.78,0.48);
  \fill[icf] (0.33,0.34) circle (0.045);
  \fill[icf] (0.78,0.48) circle (0.045);
  \draw[ic] (0.78,0.48) -- (0.99,0.32);
  \draw[ic] (0.99,0.32) -- (1.02,0.54);
  \draw[ic] (0.99,0.32) -- (1.10,0.20);}
% Operator: head and shoulders, two strokes. Reads at 6mm, which is what a
% third glyph on a client card has to survive.
\newcommand{\icperson}{%
  \draw[ic] (0.50,0.58) circle (0.155);
  \draw[ic] (0.17,0.06) .. controls (0.17,0.38) and (0.83,0.38) .. (0.83,0.06);}
\newcommand{\icchip}{%
  \draw[ic, rounded corners=0.8pt] (0.22,0.16) rectangle (0.78,0.66);
  \draw[ic] (0.34,0.28) rectangle (0.66,0.54);
  \foreach \y in {0.26,0.41,0.56}{
    \draw[ic] (0.10,\y) -- (0.22,\y);
    \draw[ic] (0.78,\y) -- (0.90,\y);}}

% ===== ROW 1 : the round loop, horizontal ===================================
% Both panels are 1.85 tall, down from 2.50: `Clients` held three cards and two
% lines of text in 2.5cm and the air was doing nothing.
%
% The strategy chips STRADDLE the controller's bottom-right corner on purpose,
% overhanging both to the right and below, each with its own shadow -- stickers
% applied to the block rather than contents that failed to fit. An overhang on
% one edge only reads as an accident, which is what the first version did.
%
% --- controller -------------------------------------------------------------
\fill[soft]  (-0.07,3.58) rectangle (4.93,5.43);
\draw[panelsrv] (0,3.65)  rectangle (5.0,5.50);
\begin{scope}[shift={(0.28,4.68)}, scale=0.76]\icserver\end{scope}
\node[ttl] at (1.30,5.22) {\textbf{Controller}};
\node[sml] at (1.30,4.86) {one round loop};
% Accent sits under the title block, not on the bottom edge: the bottom edge
% is where the stickers cross, and a rule there would be half-buried.
\draw[accsrv] (0.22,4.58) -- (4.78,4.58);
% Label sits ABOVE the sticker row, not in line with it. In line put the
% descenders of 'gy' straight through the panel's bottom edge once the
% stickers moved down to straddle it.
\node[sml, anchor=west] at (0.26,4.15) {Strategy:};
% y 3.42--3.97 against a panel bottom of 3.65, and the last chip runs to 5.48
% against a panel edge of 5.00: every sticker clears the corner on BOTH edges.
\draw[lead, draw=centcol!70!black] (1.45,4.15) -- (4.85,4.15);
\foreach \c in {2.13, 3.49, 4.85} \draw[lead, draw=centcol!70!black] (\c,4.15) -- (\c,3.99);
\foreach \x/\s in {1.50/FedAvg, 2.86/FedAdam, 4.22/FedProx}
  {\fill[soft]  ({\x-0.07},3.35) rectangle ({\x+1.19},3.90);
   \draw[chip, fill=centcol!11] (\x,3.42) rectangle ({\x+1.26},3.97);
   % accsrv, not accfed: these are the server's aggregation rules and they
   % hang off the controller panel, which is drawn in the server hue.
   \draw[accsrv] ({\x+0.10},3.48) -- ({\x+1.16},3.48);
   \node[tny] at ({\x+0.63},3.73) {\s};}

% --- the two arrows ---------------------------------------------------------
\draw[flow] (5.55,4.92) -- (6.95,4.92);
\node[tny, anchor=south] at (6.25,5.03) {model $\theta$};
\draw[flow] (6.95,4.34) -- (5.55,4.34);
\node[tny, anchor=north] at (6.25,4.23) {$\Delta\theta$, $n$};

% --- clients ----------------------------------------------------------------
\fill[soft]  (7.13,3.58) rectangle (16.43,5.43);
\draw[panelfed] (7.2,3.65) rectangle (16.5,5.50);
\node[ttl] at (7.45,5.18) {\textbf{Clients}};
\node[sml] at (8.90,5.20) {train on local data; only updates leave a client};
\draw[accfed] (7.42,5.00) -- (16.28,5.00);
% Client cards only. Each is one robot and one GPU -- the two icons say that,
% so the four repeated "robot + GPU" captions the first draft carried are gone.
\foreach \cx/\arm in {7.50/\icarmA, 10.22/\icarmB, 12.94/\icarmC} {
  \draw[sub, fill=flcol!9] (\cx,3.94) rectangle ({\cx+2.52},4.88);
  \begin{scope}[shift={({\cx+0.16},4.10)}, scale=0.62]\arm\end{scope}
  \begin{scope}[shift={({\cx+0.96},4.14)}, scale=0.58]\icperson\end{scope}
  \begin{scope}[shift={({\cx+1.66},4.10)}, scale=0.62]\icchip\end{scope}}
\node[tny, anchor=center, text=stageink!70] at (16.05,4.41) {$\cdots$};
% The axes the differing cards stand in for are NAMED IN THE CAPTION, not
% here: a fourth text line inside this panel would put back the height the
% panel was shortened to remove, and the glyphs already carry the signal.

% ===== the link between the two rows ========================================
\draw[flow] (11.85,3.58) -- (11.85,2.98);
\node[sml, anchor=west] at (12.00,3.24) {each client runs on one pair};

% ===== ROW 2 : the two pluggable axes =======================================
\fill[soft]  (-0.07,-0.07) rectangle (16.43,2.83);
\draw[panel] (0,0)         rectangle (16.5,2.90);
\node[ttl] at (0.25,2.60) {\textbf{Pluggable backends}};
\node[sml] at (3.45,2.60) {two independent axes, both set in the config file of Fig.~\ref{fig:api}};

% --- policy axis ------------------------------------------------------------
% Panel bottom raised 0.92 -> 1.12 so the chips can straddle it the way the
% Strategy stickers straddle the controller's, and still clear the
% Communicator strip below.
\draw[sub, fill=polcol!7] (0.25,1.12) rectangle (8.05,2.28);
\begin{scope}[shift={(0.45,1.50)}, scale=0.80]\icnet\end{scope}
\node[ttl] at (1.55,1.98) {Policy backend};
\node[sml] at (1.55,1.62) {VLA model:};
% Chips sit to the RIGHT within the panel, 0.20 clear of its right edge,
% rather than left-aligned under the label with dead space beside them.
\draw[lead, draw=polcol!65!black] (3.13,1.62) -- (7.025,1.62);
\foreach \c in {3.425, 5.225, 7.025} \draw[lead, draw=polcol!65!black] (\c,1.62) -- (\c,1.46);
\foreach \x/\s in {2.60/openpi, 4.40/lerobot, 6.20/groot} {
  \fill[soft] ({\x-0.07},0.82) rectangle ({\x+1.58},1.37);
  \draw[chip, fill=polcol!11] (\x,0.89) rectangle ({\x+1.65},1.44);
  \node[tny] at ({\x+0.825},1.165) {\texttt{\s}};}

% --- execution axis ---------------------------------------------------------
\draw[sub, fill=execol!7] (8.45,1.12) rectangle (16.25,2.28);
\begin{scope}[shift={(8.65,1.50)}, scale=0.80]\icscreen\end{scope}
\node[ttl] at (9.75,1.98) {Execution backend};
\node[sml] at (9.75,1.62) {FL runtime:};
\draw[lead, draw=execol!65!black] (11.27,1.62) -- (15.225,1.62);
\foreach \c in {11.625, 13.425, 15.225} \draw[lead, draw=execol!65!black] (\c,1.62) -- (\c,1.46);
\foreach \x/\s in {10.80/local, 12.60/flower, 14.40/nvflare} {
  \fill[soft] ({\x-0.07},0.82) rectangle ({\x+1.58},1.37);
  \draw[chip, fill=execol!11] (\x,0.89) rectangle ({\x+1.65},1.44);
  \node[tny] at ({\x+0.825},1.165) {\texttt{\s}};}

% --- communicator strip -----------------------------------------------------
% The strip is the layer the two axes above plug into, but as a \tiny run-on
% sentence in a dashed box it read as a footnote and left its relationship to
% the rest of the figure unstated. It now names itself in the same \ttl the
% two axes use, and opens by saying what it is to them; the description
% follows at \sml, one step up from the old \tiny.
\draw[site] (0.25,0.10) rectangle (16.25,0.74);
\node[ttl] at (0.45,0.42) {\textsc{Communicator}};
\node[sml] at (2.72,0.42)
  {both axes plug in here: the same round loop runs in-process or over a
   distributed runtime, on one cluster or across sites};

\end{tikzpicture}

%% file: sections/framework.tex
\section{The \decentvla{} Testbed}\label{sec:framework}

\decentvla{} has two parts: the \emph{framework}, a reusable
instrument, and the \emph{reference benchmark}, one configuration of
that instrument in which the partition, budgets, and evaluation are
fixed once and shared by every result of this study.

\subsection{The Framework: A Reusable Instrument}

% --- single-config figure (replaces the old plugin-API listing) -----------
% Placed before the paragraph that first references it, so the float
% lands on (or before) the page of the first reference.
\begin{figure}[tb]
\centering
\input{figures/api.tikz}
\caption{One experiment is one YAML file (abridged): the configuration
picks the policy, runtime, aggregation algorithm, partition, budgets,
and evaluation, and \texttt{decent-vla run} launches it. }
\label{fig:api}
\end{figure}

The framework sets up every experiment from a single YAML file that
\texttt{decent-vla run} executes (Fig.~\ref{fig:api}): the
configuration picks the policy, the FL runtime, the aggregation
algorithm, and all related hyperparameters. Each choice is therefore
an independent variable that can be changed one at a time, and a new
experiment, or an exact replication of a reported one, can be easily
configured. One controller drives the round loop
for every such combination (Fig.~\ref{fig:arch}). This controller contains
no policy- or runtime-specific code: \pifive{}, SmolVLA, and GR00T reach
the loop through policy adapters, and the in-process runner, Flower, and
NVFlare through runtime adapters. A client sends only the weight updates over the
federated scope of Sec.~\ref{sec:problem} and its sample count, never
raw data, and the server casts the \texttt{bf16} deltas into an
\texttt{fp32} master copy so that mixed-precision training stays
stable over many rounds. Aggregation follows the scope the policy
declares, so the same code federates a full model, a module subset,
or a LoRA update. The client partition is derived from a seed in the
configuration, so every reported run can be relaunched on the same
split.

%% file: figures/api.tikz
\definecolor{shadeink}{HTML}{16191F}

\begin{lrbox}{\apibox}
\begin{minipage}{8.06cm}
\begin{lstlisting}[style=yamlcfg]
(*\ck{name}\ckp{:}*) (*\ckv{smolvla\_libero\_fl}*)  # one experiment = one file
(*\ck{backend}\ckp{:}*)                 # policy axis: which VLA
  (*\ck{name}\ckp{:}*) (*\ckv{lerobot}*)          #   smolvla | pi05 | groot ...
  (*\ck{lora}\ckp{:}*) (*\ckp{\char`\{}\ck{r}\ckp{:}*) (*\ckn{64}\ckp{,}*) (*\ck{alpha}\ckp{:}*) (*\ckn{128}\ckp{\char`\}}*)
(*\ck{data}\ckp{:}*)
  (*\ck{repo\_id}\ckp{:}*) (*\ckv{HuggingFaceVLA/libero}*)
  (*\ck{partitioning}\ckp{:}*)
    (*\ck{scheme}\ckp{:}*) (*\ckv{task\_round\_robin}*)  # deterministic split
    (*\ck{num\_clients}\ckp{:}*) (*\ckn{8}*)
(*\ck{fl}\ckp{:}*)
  (*\ck{execution}\ckp{:}*) (*\ckv{flower}*)      # local | flower | nvflare
  (*\ck{strategy}\ckp{:}*) (*\ckp{\char`\{}\ck{name}\ckp{:}*) (*\ckv{fedavg}\ckp{\char`\}}*)  # | fedadam | fedprox
  (*\ck{rounds}\ckp{:}*) (*\ckn{100}*)
  (*\ck{local\_steps}\ckp{:}*) (*\ckn{250}*)
(*\ck{eval}\ckp{:}*)
  (*\ck{benchmark}\ckp{:}*) (*\ckp{\char`\{}\ck{name}\ckp{:}*) (*\ckv{libero}\ckp{\char`\}}*)
  (*\ck{n\_episodes\_per\_task}\ckp{:}*) (*\ckn{10}*)
\end{lstlisting}
\end{minipage}
\end{lrbox}
\setlength{\apih}{\dimexpr\ht\apibox+\dp\apibox+0.44cm\relax}

\begin{tikzpicture}[x=1cm, y=1cm,
  soft/.style={fill=shadeink, fill opacity=0.42, draw=none, rounded corners=1.5pt},
  mount/.style={fill=white, draw=shadeink, line width=0.45pt, rounded corners=1.5pt},
]
% The shadow is the mount's own rectangle translated, drawn in a shifted scope
% rather than with offset literals, so it stays correct when \apih changes with
% the listing.
\begin{scope}[shift={(-0.07,-0.07)}]
  \fill[soft] (0,0) rectangle (8.5,\apih);
\end{scope}
\draw[mount] (0,0)     rectangle (8.5,\apih);
\node[anchor=south west, inner sep=0] at (0.22,0.22) {\usebox{\apibox}};
\end{tikzpicture}

%% file: sections/benchmark_design.tex
\subsection{The Reference Benchmark}\label{sec:benchmark}
\begin{figure}[t]
\centering
\input{figures/benchmark.tikz}
\caption{The reference benchmark at a glance: the 40 tasks of the four
LIBERO suites are assigned to $K{=}8$ clients by a deterministic task
round-robin (five disjoint tasks per client), each round exchanges only
parameter updates over the federated scope $\mathcal{S}$, and every
reported score evaluates the full 40-task suite, including the tasks a
client never saw. }
\label{fig:benchmark}
\end{figure}
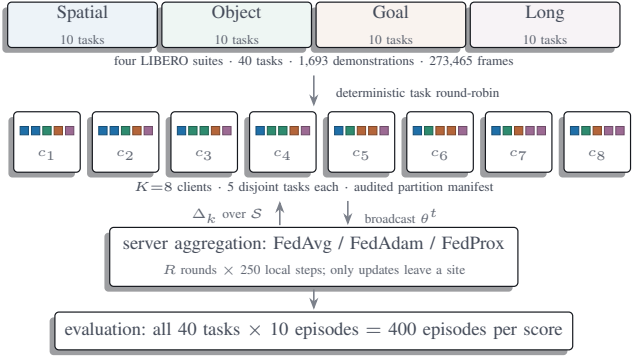

This subsection describes the benchmark configuration behind every result
of Sec.~\ref{sec:experiments} (Fig.~\ref{fig:benchmark}), with a brief
rationale for each choice. 

\subsubsection{Policies}
We benchmark three modern VLAs that span the small-to-mid-scale range
and three architectural families: \pifive{} (3.2\,B, PaliGemma),
SmolVLA (0.45\,B, SmolVLM2), and GR00T~N1.7 (3.1\,B, Cosmos-Reason2
with a DiT action head). For each family we take the strongest
openly released checkpoint; for the $\pi$-series that is
\pifive{}, as $\pi_{0.7}$~\cite{pi07} is not public. Following each
policy's published fine-tuning recipe, we fully fine-tune \pifive{}
and SmolVLA and train only GR00T's DiT action head; LoRA and
client-personalized variants are kept as controls that isolate the
scope effect (Sec.~\ref{sec:q-scope}).

\subsubsection{Data and partition}
Our simulated experiments use LIBERO~\cite{libero}, as it provides
both sufficient demonstration data and a standard evaluation
environment: 40 tasks across four suites (Spatial, Object, Goal, and Long) that
share one robot and observation space, with 1{,}693 demonstrations
and 273{,}465 frames in total. We use $K{=}8$ clients, each of which receives
five disjoint tasks by a deterministic round-robin (203--225 episodes per client). 

Our real-world experiments use six cube-stacking tasks on SO-101
arms: each task is one instruction of the form ``put the \{X\} cube on
top of the \{Y\} cube'', where \{X\} and \{Y\} name an ordered pair of
cube colors. In each experiment, every task has 50 demonstrations, 300 in
total, all collected at the site that owns the task. In the two-client experiment each
client trains on the three tasks it collected; in the three-client
experiment each client trains on the two stacking orders of one color
pair (Sec.~\ref{sec:q-real}). We choose the SO-101 here because federated
learning presupposes many potential participating sites, and this
open-source, low-cost robot arm is one of the most popular: WowRobo
Robotics, one primary vendor of the kits, reports over 10{,}000 SO-101 units
shipped worldwide~\cite{wowrobo}.

\subsubsection{Training setup}
Unless stated otherwise the server aggregates with FedAvg (FedAdam with
server learning rate $10^{-3}$ and FedProx with $\mu{=}0.01$ are
compared in Sec.~\ref{sec:q-rounds}), and clients run 250
local steps per round with AdamW at each policy's published
fine-tuning hyperparameters, so that the benchmark measures the
effect of federating
rather than of hyperparameter tuning. LoRA controls use rank 64 with
$\alpha=128$ on the attention and MLP projections, applied to backbone
and action head for \pifive{} and SmolVLA; for GR00T the adapters
attach to the DiT action head, with the backbone frozen and the
embodiment projector trained full-rank. Each simulated result is reported as the run's best score over the
rounds evaluated (Fig.~\ref{fig:convergence}); the aggregation
comparison of Sec.~\ref{sec:q-rounds} and the personalization runs use
a matched 100-round budget, and the robustness evaluation of
Sec.~\ref{sec:q-robust} rescores the checkpoints of Table~\ref{tab:main}. The centralized
references are fine-tuned to convergence on the pooled 40-task data
(SmolVLA through \decentvla{} with one client, GR00T through its
official fine-tuning script) or, for \pifive{}, taken from the publicly
released LIBERO checkpoint~\cite{pi05} and rescored in our harness at
50 episodes per task.

\subsubsection{Evaluation in simulation}
We evaluate the centralized and federated fine-tuned models on the full 40
tasks in LIBERO with 10 trials each and report the success rate averaged over
the 400 episodes; SmolVLA and \pifive{} execute 10-step action chunks
between replans; GR00T replans every 16 steps. The personalization runs of Sec.~\ref{sec:q-scope} additionally
evaluate the clients: each client's own model is scored on the five
tasks it owns, again at 10 episodes per task, so the eight clients
together still yield 400 episodes.

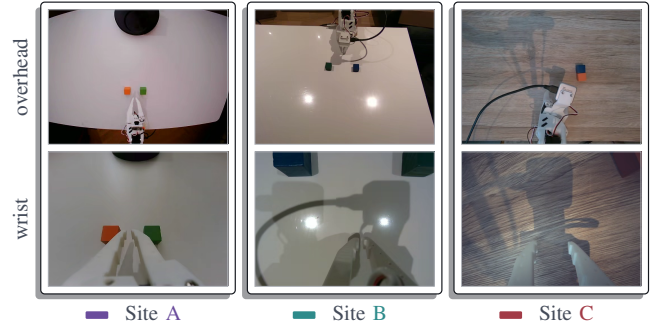
\begin{figure}[t]
\centering
\input{figures/sites.tikz}
\caption{Three physical sites (A--C): the overhead and wrist camera
views that form the policy's input. Heterogeneity is both visual and
physical, the latter from differing hardware calibration
(Fig.~\ref{fig:calib}).}
\label{fig:sites}
\end{figure}

\subsubsection{Evaluation in the real world}
We evaluate each model on all six tasks at each site, with 10 trials
per task judged pass/fail by the operator from the rollout, and report
the success rate averaged over the 60 episodes. The two experiments
train separate models and are not compared to each other
(Sec.~\ref{sec:q-real}).

\subsubsection{Real-to-simulation evaluation}\label{sec:realsim-eval}
\begin{figure}[t]
\centering
\input{figures/simstages.tikz}
\caption{The simulated environment of SO-101. \textbf{Left:} the overhead and wrist views
the policy receives. \textbf{Right:} two rollouts of one \pifive{}
checkpoint, one completing the stack and one not, with each stage shown
as a composite of the five frames before it triggers.}
\label{fig:realsim}
\end{figure}
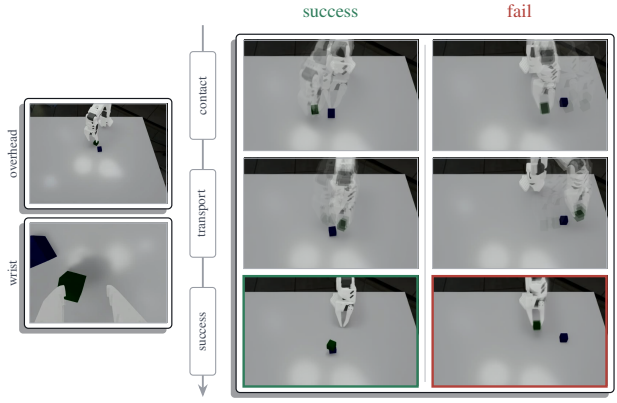

Real-world trials are costly and cannot control object position and pose, so we additionally rescore the three-client checkpoints
in a simulated environment (Fig.~\ref{fig:realsim}), which affords repeatable layouts, automated evaluation at scale, and finer
stage-wise success rates. The simulated environment is built in Isaac Sim~\cite{isaacsim} with RoboLab~\cite{robolab},
which provides a photorealistic environment for the SO-101 and objects.

We calibrate the simulation with real-world cube positions: for each
task, we read where the cubes were at the moment the gripper closed
in its demonstrations. This gives 258 layouts that the models were
trained on, which serve as the unperturbed 0\,cm control. We then
shift the cubes by up to 1, 3, or 5\,cm from these positions and
evaluate the centralized and federated \pifive{} at each amplitude.
Layouts whose cubes would overlap are redrawn.

The stage-wise success rate scores each rollout by the furthest of
three stages it reaches: \emph{contact}, any gripper--cube contact;
\emph{transport}, the cube carried in the air over the target; and
\emph{success}, the stack completed and holding.

%% file: figures/benchmark.tikz
\definecolor{suiteS}{HTML}{1F6FA8}   % Spatial -- blue
\definecolor{suiteO}{HTML}{2E8B6F}   % Object  -- bluish green
\definecolor{suiteG}{HTML}{B4653A}   % Goal    -- vermillion, muted
\definecolor{suiteL}{HTML}{9C6A93}   % Long    -- reddish purple, muted
\definecolor{shadeink}{HTML}{16191F}
\definecolor{stageink}{HTML}{444B59}
\begin{tikzpicture}[
  x=1cm, y=1cm,
  soft/.style={fill=shadeink, fill opacity=0.42, draw=none, rounded corners=1.5pt},
  suitebox/.style={draw=shadeink, fill=#1!8, rounded corners=1.5pt,
      line width=0.45pt, minimum width=1.98cm, minimum height=0.60cm,
      align=center, inner sep=1pt},
  clientcard/.style={draw=shadeink, fill=white, rounded corners=1.5pt,
      line width=0.45pt, minimum width=0.88cm, minimum height=0.80cm,
      inner sep=0pt},
  mountbox/.style={draw=shadeink, fill=white, rounded corners=1.5pt,
      line width=0.45pt, align=center},
  task/.style={draw=#1!70!black, fill=#1, line width=0.3pt},
  flow/.style={stageink!70, line width=0.6pt,
      -{Stealth[length=1.6mm, width=1.2mm]}},
  lbl/.style={font=\tiny, text=stageink},
  sub/.style={font=\tiny, text=stageink!85},
]

% ---- suites --------------------------------------------------------------
\foreach \x/\c/\name in {1.19/suiteS/Spatial, 3.23/suiteO/Object,
                         5.27/suiteG/Goal,    7.31/suiteL/Long}{
  \fill[soft] (\x-1.06,4.68) rectangle (\x+0.92,5.28);
  \node[suitebox=\c] at (\x,5.05)
    {\scriptsize\textcolor{stageink}{\name}\\[-2.5pt]\tiny\textcolor{stageink!85}{10 tasks}};
}
\node[lbl] at (4.25,4.56)
  {four LIBERO suites $\cdot$ 40 tasks $\cdot$ 1{,}693 demonstrations $\cdot$ 273{,}465 frames};

% ---- partition arrow -----------------------------------------------------
\draw[flow] (4.25,4.38) -- (4.25,3.97);
\node[lbl, anchor=west] at (4.42,4.15) {deterministic task round-robin};

% ---- client cards (5 tasks each; colors = true round-robin composition) --
% client k (1-indexed) owns tasks {k-1, k+7, k+15, k+23, k+31} of the
% suite-concatenated task list, hence the suite mix below.
\foreach \i/\ca/\cb/\cc/\cd/\ce in {%
  0/suiteS/suiteS/suiteO/suiteG/suiteL,
  1/suiteS/suiteS/suiteO/suiteG/suiteL,
  2/suiteS/suiteO/suiteO/suiteG/suiteL,
  3/suiteS/suiteO/suiteO/suiteG/suiteL,
  4/suiteS/suiteO/suiteG/suiteG/suiteL,
  5/suiteS/suiteO/suiteG/suiteG/suiteL,
  6/suiteS/suiteO/suiteG/suiteL/suiteL,
  7/suiteS/suiteO/suiteG/suiteL/suiteL}{
  \pgfmathsetmacro{\cx}{0.72 + \i*1.04}
  \fill[soft] (\cx-0.51,3.03) rectangle (\cx+0.37,3.83);
  \node[clientcard] at (\cx,3.50) {};
  \foreach \j/\tc in {0/\ca,1/\cb,2/\cc,3/\cd,4/\ce}{
    \pgfmathsetmacro{\sx}{\cx - 0.30 + \j*0.15}
    \draw[task=\tc] (\sx-0.055,3.595) rectangle (\sx+0.055,3.705);
  }
  \pgfmathtruncatemacro{\k}{\i+1}
  \node[sub] at (\cx,3.33) {$c_{\k}$};
}
% The card row was lifted 0.05 and this label dropped 0.03 from their
% pre-restyle positions: the cards now throw a 0.07cm shadow below their
% lower edge, and the label used to sit flush against that edge.
\node[lbl] at (4.25,2.89)
  {$K{=}8$ clients $\cdot$ 5 disjoint tasks each $\cdot$ audited partition manifest};

% ---- round loop ----------------------------------------------------------
% Both arrows are neutral: the up-arrow used to be drawn and labelled in the
% Spatial suite's blue, which put a series colour on text and implied a tie
% to that suite that does not exist.
\draw[flow] (3.80,2.40) -- (3.80,2.72);
\node[sub, anchor=east] at (3.70,2.52) {$\Delta_k$ over $\mathcal{S}$};
\draw[flow] (4.70,2.72) -- (4.70,2.40);
\node[sub, anchor=west] at (4.80,2.52) {broadcast $\theta^{t}$};
% minimum height must EXCEED the content (0.77cm) or it stops binding
% and these hand-computed shadow corners no longer match the box.
\fill[soft] (1.48,1.52) rectangle (6.88,2.30);
\node[mountbox, minimum width=5.4cm, minimum height=0.78cm] at (4.25,1.98)
  {\scriptsize\textcolor{stageink}{server aggregation: FedAvg / FedAdam / FedProx}\\[-2.5pt]
   \tiny\textcolor{stageink!85}{$R$ rounds $\times$ 250 local steps; only updates leave a site}};

% ---- evaluation ----------------------------------------------------------
\draw[flow] (4.25,1.55) -- (4.25,1.28);
\fill[soft] (0.755,0.68) rectangle (7.605,1.18);
\node[mountbox, minimum width=6.85cm, minimum height=0.50cm] at (4.25,1.00)
  {\scriptsize\textcolor{stageink}{evaluation: all 40 tasks $\times$ 10 episodes $=$ 400 episodes per score}};

\end{tikzpicture}

%% file: figures/sites.tikz
\definecolor{shadeink}{HTML}{16191F}
\definecolor{stageink}{HTML}{444B59}
\definecolor{siteAcol}{HTML}{6A4C9C}
\definecolor{siteBcol}{HTML}{2E8B8B}
\definecolor{siteCcol}{HTML}{A23B47}

\begin{tikzpicture}[x=1cm, y=1cm,
  soft/.style={fill=shadeink, fill opacity=0.42, draw=none, rounded corners=1.5pt},
  mount/.style={fill=white, draw=shadeink, line width=0.45pt, rounded corners=1.5pt},
  frame/.style={draw=stageink!45, line width=0.3pt},
]

% ---- one box per site ------------------------------------------------------
% \s is the file stem, \x the box's left edge. Overhead sits above wrist, so
% the pair reads top-down exactly as the policy's two streams are listed
% everywhere else in the paper.
\foreach \s/\x in {siteA/0.46, siteB/3.1933, siteC/5.9266} {
  \fill[soft] ({\x-0.07},-0.07) rectangle ({\x+2.5033},3.79);
  \draw[mount] (\x,0) rectangle ({\x+2.5733},3.86);
  \node[anchor=south west, inner sep=0] at ({\x+0.10},1.98)
    {\includegraphics[width=2.3733cm, height=1.78cm]{figures/photos/real/\s-front.png}};
  \draw[frame] ({\x+0.10},1.98) rectangle ({\x+2.4733},3.76);
  \node[anchor=south west, inner sep=0] at ({\x+0.10},0.10)
    {\includegraphics[width=2.3733cm, height=1.78cm]{figures/photos/real/\s-wrist.png}};
  \draw[frame] ({\x+0.10},0.10) rectangle ({\x+2.4733},1.88);
}

% ---- row labels ------------------------------------------------------------
% Rotated into the 0.40cm gutter, centred on each row of views. These label
% rows that run across all three boxes, which is why they sit outside the
% boxes rather than being repeated inside each one.
\foreach \y/\s in {2.87/overhead, 0.99/wrist}
  \node[rotate=90, anchor=center, font=\footnotesize, text=stageink]
    at (0.19,\y) {\s};

% ---- site labels -----------------------------------------------------------
% anchor=base at -0.34, the drop the stage names use in stage.tikz.
% The label sits 0.20 right of the box centre so that swatch + gap + text is
% what ends up centred under the box, not the text alone. Swatch is 0.30 wide
% and 0.10 tall, occupying the x-height band so it reads as part of the line.
\foreach \x/\s/\c in {1.7467/{Site \textcolor{siteAcol}{A}}/siteAcol,
                      4.48/{Site \textcolor{siteBcol}{B}}/siteBcol,
                      7.2133/{Site \textcolor{siteCcol}{C}}/siteCcol} {
  \node[anchor=base, font=\footnotesize, text=stageink] (lbl\c)
    at ({\x+0.20},-0.34) {\s};
  \fill[\c, rounded corners=0.5pt]
    ([xshift=-0.40cm, yshift=0.01cm]lbl\c.base west) rectangle
    ([xshift=-0.10cm, yshift=0.11cm]lbl\c.base west);
}

\end{tikzpicture}

%% file: figures/simstages.tikz
\definecolor{shadeink}{HTML}{16191F}
\definecolor{stageink}{HTML}{444B59}
\definecolor{okcol}{HTML}{2E7D5B}   % success -- status colour, always labelled
\definecolor{nocol}{HTML}{B3413A}   % fail    -- status colour, always labelled

\begin{tikzpicture}[x=1cm, y=1cm,
  soft/.style={fill=shadeink, fill opacity=0.42, draw=none, rounded corners=1.5pt},
  mount/.style={fill=white, draw=shadeink, line width=0.45pt, rounded corners=1.5pt},
  frame/.style={draw=stageink!45, line width=0.3pt},
  lbl/.style={font=\tiny, text=stageink},
  tag/.style={draw=stageink!55, fill=white, line width=0.4pt, rounded corners=1.2pt,
              inner sep=0pt, minimum width=0.34cm, minimum height=1.16cm},
  flow/.style={stageink!55, line width=0.7pt, -{Stealth[length=1.7mm, width=1.3mm]}},
]

% ===== left: the two views the policy receives ==============================
% Deliberately the smallest elements here: they say "this is the scene", the
% argument is on the right.
\foreach \y/\img/\lab in {2.43/view-overhead/overhead, 0.85/view-wrist/wrist} {
  \fill[soft]  (0.33,{\y-0.07}) rectangle (2.28,{\y+1.39});
  \draw[mount] (0.40,\y)        rectangle (2.35,{\y+1.46});
  \node[anchor=south west, inner sep=0] at (0.47,{\y+0.07})
    {\includegraphics[width=1.81cm, height=1.32cm]{figures/photos/sim/\img.png}};
  \draw[frame] (0.47,{\y+0.07}) rectangle (2.28,{\y+1.39});
  \node[rotate=90, anchor=center, lbl] at (0.24,{\y+0.73}) {\lab};}

% ===== the progression arrow, drawn FIRST so the tags sit on top of it ======
\draw[flow] (2.76,4.86) -- (2.76,-0.06);

% ===== right: one box, two columns, three rows ==============================
\fill[soft]  (3.13,-0.07) rectangle (8.13,4.67);
\draw[mount] (3.20,0)     rectangle (8.20,4.74);
% rule between the columns rather than a second box: the two rollouts are the
% same experiment, not two experiments.
\draw[stageink!35, line width=0.4pt] (5.70,0.14) -- (5.70,4.60);
% \y = row bottom; \a success-column file; \b fail-column file
\foreach \y/\a/\b in {3.20/ok-contact/no-contact,
                      1.64/ok-transport/no-transport,
                      0.08/ok-success/no-nostack} {
  \node[anchor=south west, inner sep=0] at (3.30,\y)
    {\includegraphics[width=2.30cm, height=1.46cm]{figures/photos/sim/\a.png}};
  \draw[frame] (3.30,\y) rectangle (5.60,{\y+1.46});
  \node[anchor=south west, inner sep=0] at (5.80,\y)
    {\includegraphics[width=2.30cm, height=1.46cm]{figures/photos/sim/\b.png}};
  \draw[frame] (5.80,\y) rectangle (8.10,{\y+1.46});}

% ===== the outcome, ringed on the row that decides it =======================
\draw[okcol, line width=0.9pt] (3.30,0.08) rectangle (5.60,1.54);
\draw[nocol, line width=0.9pt] (5.80,0.08) rectangle (8.10,1.54);
\node[font=\scriptsize, text=okcol, anchor=south] at (4.45,4.84) {success};
\node[font=\scriptsize, text=nocol, anchor=south] at (6.95,4.84) {fail};

% ===== stage tags, on top of the arrow ======================================
\foreach \y/\s in {3.93/contact, 2.37/transport, 0.81/success}
  \node[tag] at (2.76,\y) {\rotatebox{90}{\tiny\textcolor{stageink}{\s}}};

\end{tikzpicture}

%% file: sections/experiments.tex
\section{Results}\label{sec:experiments}

Each subsection below answers one question of Sec.~\ref{sec:problem} by
varying the factor under study on the reference benchmark of
Sec.~\ref{sec:benchmark}. Success rates and their differences are reported
in absolute percentage terms.

% ================================================================= Q1
\begin{table}[t]
\caption{Federated versus centralized fine-tuning: LIBERO success rate
(\%); best per column in bold.}
\label{tab:main}
\centering
\footnotesize
\renewcommand{\arraystretch}{1.15}
\setlength{\tabcolsep}{3.0pt}
\begin{tabular}{l ccccc}
\toprule
Training & Spatial & Object & Goal & Long & Overall \\
\midrule
\multicolumn{6}{l}{{\boldmath\pifive{}} (3.2\,B)~\cite{pi05}} \\
\quad Centralized & \textbf{97.0} & 99.0 & 97.0 & 95.0 & 97.00 \\
\quad Federated   & 92.0 & \textbf{100.0} & \textbf{99.0} & \textbf{99.0} & \textbf{97.50} \\
\midrule
\multicolumn{6}{l}{\textbf{GR00T~N1.7} (3.1\,B)~\cite{grootn17}} \\
\quad Centralized & \textbf{97.0} & 97.0  & \textbf{99.0} & \textbf{94.0} & \textbf{96.75} \\
\quad Federated   & 91.0 & \textbf{99.0} & 95.0 & 79.0 & 91.00 \\
\midrule
\multicolumn{6}{l}{\textbf{SmolVLA} (0.45\,B)~\cite{smolvla}} \\
\quad Centralized & \textbf{89.0} & \textbf{97.0}  & \textbf{95.0} & \textbf{79.0} & \textbf{90.00} \\
\quad Federated   & 75.0 & 70.0  & 81.0 & 75.0 & 75.25 \\
\bottomrule
\end{tabular}
\end{table}

\subsection{Federated Versus Centralized Fine-Tuning (Q1)}\label{sec:q-gap}

\textbf{Federated fine-tuning matches or approaches centralized
fine-tuning for the two 3\,B policies.} In Table~\ref{tab:main},
federated fine-tuned \pifive{} achieves 97.50\% against its 97.00\% centralized
reference, and GR00T~N1.7 reaches 91.00\% against 96.75\%, a 5.75\%
gap that concentrates in Long, whose 15.0\% drop supplies two thirds
of it. Federated fine-tuning is therefore promising at this scale:
no demonstration leaves its client, yet both policies stay within
5.75\% of their centralized references, and \pifive{} exceeds its
reference.

SmolVLA, in contrast, finishes 14.75\% below its centralized 
reference (75.25\% against 90.00\%). The deficit is largest on Object (97.0\% against 70.0\%) and
smallest on Long (79.0\% against 75.0\%). FedAvg differs from centralized
fine-tuning in exactly one step: each round it averages the eight
client models. Averaging works only while the clients stay close to
the shared pretrained model~\cite{modelsoups,fedpretrain,wheretobegin}, so the
strength of the pretraining decides the outcome. \pifive{} builds on
over 10{,}000 hours of cross-embodiment robot data~\cite{pi0,pi05},
and GR00T on real, simulated, and human-video corpora spanning
thousands of hours~\cite{groot}: fitting LIBERO needs only small
updates. SmolVLA is pretrained
on roughly 100 hours of community-collected SO-100 data (10.6M frames
at ${\sim}30$\,fps)~\cite{smolvla}, two orders of magnitude less: each
client must move far toward its own five tasks, the models drift
apart, and their average degrades~\cite{scaffold}.
% ================================================================= Q2
\begin{table}[t]
\caption{Full fine-tuning versus LoRA: payload and LIBERO success rate
(\%); best per policy in bold.}
\label{tab:scope-lora}
\centering
\footnotesize
\renewcommand{\arraystretch}{1.15}
\setlength{\tabcolsep}{4pt}
\begin{tabular}{l r r c}
\toprule
Update parameterization & Params & Payload & Success (\%) \\
\midrule
\multicolumn{4}{l}{{\boldmath\pifive{}}} \\
\quad Full fine-tuning        & 3.62\,B & 7.24\,GB & \textbf{97.50} \\
\quad LoRA                    & 0.12\,B & 0.24\,GB & \textbf{97.50} \\
\midrule
\multicolumn{4}{l}{\textbf{GR00T~N1.7}} \\
\quad Action head (full)      & 1.62\,B & 3.24\,GB & \textbf{91.00} \\
\quad Action head (LoRA)      & 0.39\,B & 0.77\,GB & 87.00 \\
\midrule
\multicolumn{4}{l}{\textbf{SmolVLA}} \\
\quad Full fine-tuning        & 0.40\,B & 0.81\,GB & \textbf{75.25} \\
\quad LoRA                    & 0.04\,B & 0.08\,GB & 55.25 \\
\bottomrule
\end{tabular}
\end{table}

\begin{table}[t]
\caption{Partial federation: personalized LIBERO success rate (\%);
best per policy in bold. Only the listed scope is federated, the
remainder stays client-personal.}
\label{tab:scope-personal}
\centering
\footnotesize
\renewcommand{\arraystretch}{1.15}
\setlength{\tabcolsep}{3.3pt}
\begin{tabular}{l r r r c}
\toprule
Federated scope & Shared & Personal & Payload & Personalized success (\%) \\
\midrule
\multicolumn{5}{l}{{\boldmath\pifive{}}} \\
\quad Action head & 0.69\,B & 2.92\,B & 1.39\,GB & 95.25 \\
\quad Backbone    & 2.92\,B & 0.69\,B & 5.85\,GB & \textbf{97.25} \\
\midrule
\multicolumn{5}{l}{\textbf{GR00T~N1.7}} \\
\quad Action head & 1.62\,B & 1.52\,B & 3.24\,GB & \textbf{86.25} \\
\quad Backbone    & 1.52\,B & 1.62\,B & 3.05\,GB & 71.25 \\
\midrule
\multicolumn{5}{l}{\textbf{SmolVLA}} \\
\quad Action head & 0.10\,B & 0.30\,B & 0.20\,GB & 81.50 \\
\quad Backbone    & 0.30\,B & 0.10\,B & 0.61\,GB & \textbf{85.75} \\
\bottomrule
\end{tabular}
\end{table}

\subsection{Effect of the Federated Scope (Q2)}\label{sec:q-scope}
Choosing $\mathcal{S}$ involves two decisions: how the update is
parameterized, by full fine-tuning or by LoRA, and which parameters are
federated at all. Tables~\ref{tab:scope-lora}
and~\ref{tab:scope-personal} ablate the two decisions separately.

\textbf{LoRA matches full fine-tuning for \pifive{} but trails it by
4.00\% for GR00T and by 20.00\% for SmolVLA.} In
Table~\ref{tab:scope-lora}, \pifive{} recovers its full fine-tuning
score exactly (97.50\%), GR00T falls from 91.00\% to 87.00\%, and
SmolVLA from 75.25\% to 55.25\%. The performance difference is consistent with the findings in Sec.~\ref{sec:q-gap}: 
the weaker a policy's pretraining, the larger the update the target demands, while rank 64 fixes how large an update a client
can express~\cite{intrinsicdim,loralearnsless}. 

\begin{figure}[t]
\centering
\input{figures/convergence.tikz}
\caption{Full fine-tuning (solid) versus LoRA (dashed) under global
federation: LIBERO success rate over the round budget, for \pifive{}
(3.2\,B), GR00T~N1.7 (3.1\,B, action head) and SmolVLA (0.45\,B).
Every marker is a full 400-episode evaluation.}
\label{fig:convergence}
\end{figure}
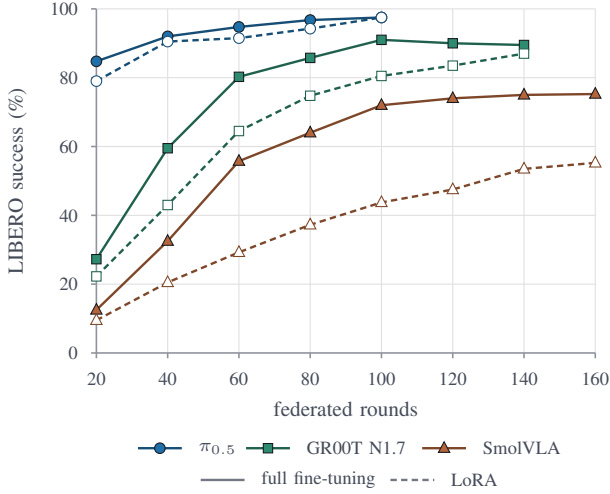

\textbf{LoRA reduces the transmission payload but converges more
slowly.} In Fig.~\ref{fig:convergence} every LoRA curve runs behind
its full fine-tuning counterpart, and the lag again follows the
pretraining strength: roughly 20 rounds for \pifive{}, which closes
the gap by round 100; 40--60 for GR00T (87.00\% at round 140 against 85.75\% at
the full curve's round 80); roughly 100 for SmolVLA, which reaches
at round 160 what its full curve passed by round 60. In exchange the
payload falls $30\times$, $4.2\times$, and $10\times$
(Table~\ref{tab:scope-lora}). Under limited bandwidth LoRA is
therefore practical for a strongly pretrained policy, paying rounds
for a smaller per-round payload.

\textbf{Personalization improves SmolVLA but not the two 3\,B
policies, and it leaves no usable global model.} In
Table~\ref{tab:scope-personal} only the listed module is federated,
the remainder stays client-personal, every client fully fine-tunes
(including GR00T's otherwise-frozen backbone), and each client is
scored on its own five tasks (\emph{personalized success}).
Personalization helps where averaging hurt: SmolVLA, whose deficit
Sec.~\ref{sec:q-gap} traced to averaging, recovers 10.50\% once its
action head stays personal (85.75\% with the backbone federated,
against its 75.25\% federated score). \pifive{} is essentially
unchanged (97.25\% against 97.50\%), and GR00T
reaches 86.25\% against its 91.00\% federated score, a difference
concentrated in two tasks of a single client. The price is the
global model: the server checkpoint holds only the shared module,
only ever trained alongside the clients' personal remainders, and
scores 0.00\% in every configuration we evaluated (not shown).
Which module to federate has no universal answer: the backbone wins
for \pifive{} and SmolVLA, the action head for GR00T.

% ================================================================= Q3
\subsection{Effect of the Aggregation Algorithm (Q3)}\label{sec:q-rounds}

\begin{table}[t]
\caption{Aggregation algorithms on \pifive{} full fine-tuning: LIBERO
success rate (\%); best per column in bold.}
\label{tab:agg}
\centering
\footnotesize
\renewcommand{\arraystretch}{1.15}
\setlength{\tabcolsep}{3.0pt}
\begin{tabular}{l ccccc}
\toprule
Algorithm & Spatial & Object & Goal & Long & Overall \\
\midrule
FedAvg~\cite{fedavg}   & \textbf{92.0} & \textbf{100.0} & \textbf{99.0} & \textbf{99.0} & \textbf{97.50} \\
FedAdam~\cite{fedopt}  & 91.0 & \textbf{100.0} & \textbf{99.0} & 95.0 & 96.25 \\
FedProx~\cite{fedprox} & 89.0 & 99.0 & 95.0 & 94.0 & 94.25 \\
\bottomrule
\end{tabular}
\end{table}

\textbf{Neither FedAdam nor FedProx improves on FedAvg under
precisely the client heterogeneity they are designed for.}
Table~\ref{tab:agg} compares the three algorithms on \pifive{} full
fine-tuning under a matched budget with everything else held fixed:
FedAvg finishes highest at 97.50\% and matches or exceeds both
alternatives on every suite, FedAdam reaches 96.25\%, and FedProx
94.25\%. Each alternative adds one mechanism to FedAvg, and neither
pays off. FedAdam adds server-side adaptivity on top of the AdamW every
client already runs locally, and costs 1.25\%. FedProx
adds a penalty on local drift from the broadcast model, but under
the task partition each client's optimum is genuinely far from the
global one, so the penalty withholds part of each round's progress
and costs 3.25\%. 

% --- Q5 floats (Figs. 8-10) are sourced here, at the end of Q3, so that
% they are encountered one page before the Q5 text and land on the Q5
% page instead of drifting after the conclusion, between the references.
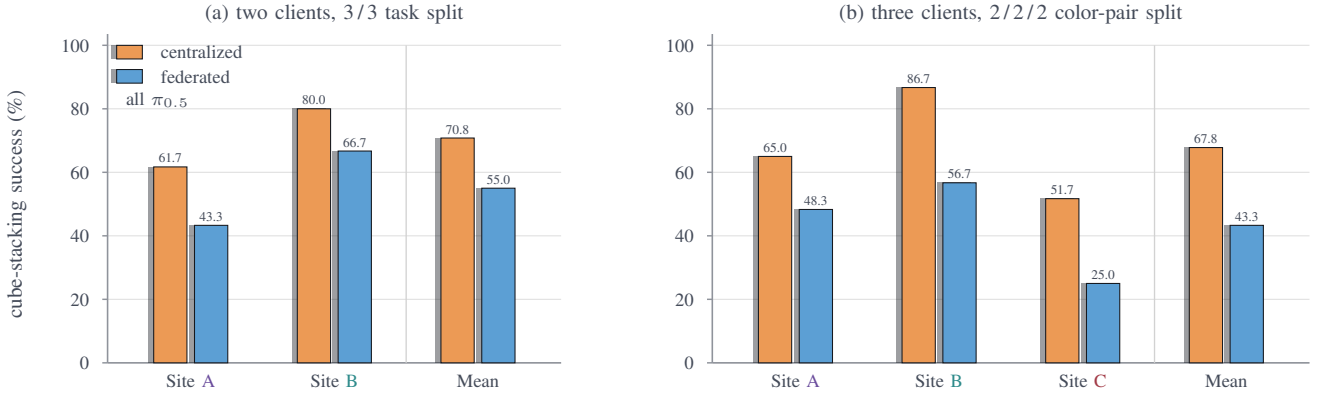
\begin{figure*}[t]
\centering
\input{figures/real.tikz}
\caption{Real-robot transfer: SO-101 cube-stacking success on each site's
arm, over all six tasks, for the centralized and the FedAvg server
\pifive{} models. \textbf{(a)}~Two clients, task split;
\textbf{(b)}~three clients, color-pair split; each bar pools 60 trials
and the per-experiment means pool 120 and 180.}
\label{fig:real}
\end{figure*}

% Both single-column Q5 floats are sourced here, right after the Q5
% heading, so they land on the Q5 pages instead of drifting past the
% conclusion.
\begin{figure}[t]
\centering
% Normalized to the column measure so the fixed-cm drawing fits both the
% IEEEtran (RA-L) and ieeeconf (ICRA) column widths.
\resizebox{\columnwidth}{!}{\input{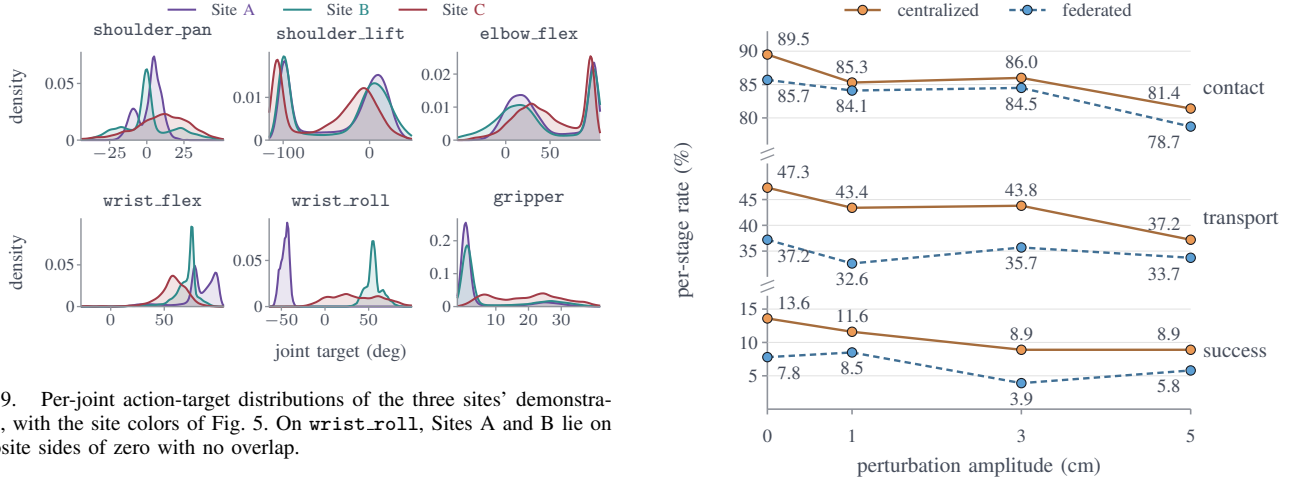}}
\caption{Per-joint action-target distributions of the three sites'
demonstrations, with the site colors of Fig.~\ref{fig:sites}. On
\texttt{wrist\_roll}, Sites A and B lie on opposite sides of zero with
no overlap.}
\label{fig:calib}
\end{figure}

\begin{figure}[t]
\centering
\input{figures/purt.tikz}
\caption{Stage-wise success rate (\%) for the centralized and
federated \pifive{} under object-pose perturbation in the simulated
environment (Fig.~\ref{fig:realsim}): 258 paired layouts per
amplitude, identical between models.}
\label{fig:purt}
\end{figure}

% ================================================================= Q4
\subsection{Robustness Under Distribution Shift (Q4)}\label{sec:q-robust}

\begin{table}[t]
\caption{Robustness to distribution shift: LIBERO-PRO success rate
(\%); best per column in bold.}
\label{tab:pro}
\centering
\footnotesize
\renewcommand{\arraystretch}{1.15}
\setlength{\tabcolsep}{3.2pt}
\begin{tabular}{l ccccc c}
\toprule
Training & Base & Language & Object & Position & Task & Overall \\
\midrule
\multicolumn{7}{l}{{\boldmath\pifive{}}} \\
\quad Centralized & 97.00 & \textbf{97.50} & \textbf{87.50} & 22.50 & 4.50 & 53.00 \\
\quad Federated   & \textbf{97.50} & 97.00 & 83.50 & \textbf{43.25} & \textbf{4.75} & \textbf{57.13} \\
\midrule
\multicolumn{7}{l}{\textbf{GR00T~N1.7}} \\
\quad Centralized & \textbf{96.75} & \textbf{93.25} & \textbf{79.00} & 2.00 & \textbf{5.00} & \textbf{44.81} \\
\quad Federated   & 91.00 & 92.25 & 73.75 & \textbf{8.75} & 4.00 & 44.69 \\
\midrule
\multicolumn{7}{l}{\textbf{SmolVLA}} \\
\quad Centralized & \textbf{90.00} & \textbf{89.50} & \textbf{62.75} & 0.00 & 3.25 & \textbf{38.88} \\
\quad Federated   & 75.25 & 77.00 & 56.50 & \textbf{0.50} & \textbf{4.00} & 34.50 \\
\bottomrule
\end{tabular}
\end{table}

\textbf{Federated fine-tuning is as robust to distribution
shift as centralized fine-tuning.} Table~\ref{tab:pro} rescores the
centralized and federated models on LIBERO-PRO~\cite{liberopro},
which re-renders each suite under instruction paraphrase
(\emph{Language}), object appearance change (\emph{Object}),
object position swap (\emph{Position}), and goal redefinition
(\emph{Task}). Every model degrades sharply, and goal redefinition
defeats all of them (${\leq}\,5.00\%$). The federated deficits,
however, shrink under shift: GR00T's 5.75\% in-distribution gap
narrows to 0.12\%, SmolVLA's 14.75\% narrows to 4.38\%, and
\pifive{} moves 4.13\% ahead of the centralized reference it
matched in distribution.

\textbf{The federated advantage concentrates in the position swap.}
Under the swap every federated row leads its centralized
counterpart: 43.25\% against 22.50\% for \pifive{}, 8.75\% against
2.00\% for GR00T, and 0.50\% against 0.00\% for SmolVLA; \pifive{}'s
20.75\% lead here contributes 5.19\% to its 4.13\% overall lead,
outweighing the centralized advantages under paraphrase and
appearance change. A candidate explanation is that federation adds
an implicit regularization. LIBERO-PRO traces the collapse under the position
swap to policies memorizing the absolute object placements of their
training layouts~\cite{liberopro}, and such memorization is what
averaging removes: each client can memorize only the layouts of its
own five tasks, and the round-wise average keeps what the eight
models share while canceling what they do
not~\cite{modelsoups,diwa}. The centralized model, trained on all 40
layouts in a single run and never averaged, is free to memorize
them.

% ================================================================= Q5
\subsection{Transfer to Physical Robots (Q5)}\label{sec:q-real}

\textbf{In real-world evaluation, federated fine-tuning performs worse
than centralized fine-tuning in both experiments.}
Fig.~\ref{fig:real} scores the two \pifive{} models of each
experiment on all six tasks at every site: with two task-split
clients the centralized model leads by 15.8\% (70.8\% against
55.0\%), with three color-pair clients by 24.5\% (67.8\% against
43.3\%). Combining the sites' demonstrations by parameter averaging
recovers less than centralizing them directly, although the same
policy's federated fine-tuning matched centralized fine-tuning in simulation
(Sec.~\ref{sec:q-gap}).

The likely cause is heterogeneity: the real sites differ in more
ways than the LIBERO partition expresses. In LIBERO, clients differ
only in which tasks they hold, and share one simulated robot. The
physical sites differ in their tasks; in their observations, through
camera placement, lighting, and background (Fig.~\ref{fig:sites});
and, above all, in their actions and states, through each arm's
calibration and each operator's habits. Fig.~\ref{fig:calib}
measures the action axis: although every arm is the same SO-101
model, the \texttt{wrist\_roll} distributions of Sites A and B sit
$101.8^\circ$ apart (Wasserstein distance), on opposite sides of
zero with no overlap, and \texttt{wrist\_flex} and
\texttt{shoulder\_pan} also shift across sites. Averaging across
this spread degrades action precision rather than task grounding: in
the recorded rollouts the federated policy typically reaches the
correct cube but cannot grasp it stably or stack it precisely.

\textbf{The real-to-simulation evaluation confirms the real-world
result: centralized fine-tuning leads at every perturbation
amplitude.} We rescore the two three-client checkpoints in the
simulated environment of Sec.~\ref{sec:realsim-eval}, on the
demonstrated cube layouts and under displacements of up to 1, 3, and
5\,cm (Fig.~\ref{fig:purt}). At 0\,cm, the in-distribution control,
centralized fine-tuning reaches \emph{transport} in 47.3\% of
episodes against 37.2\% and \emph{success} in 13.6\% against 7.8\%,
reproducing in simulation the lead the two real-world experiments of
Fig.~\ref{fig:real} found. Moving right in
Fig.~\ref{fig:purt}, both models degrade as the cubes are displaced,
and the ordering never flips: no amplitude in the tested range
brings the federated policy level with the centralized one.

%% file: figures/convergence.tikz
% Fig. (fig:convergence) --- full fine-tuning vs LoRA over the round budget.
%
% GENERATED by figures/make_convergence_tikz.py -- edit that, not this.
%
% Port of the matplotlib original, which is kept at figures/convergence.pdf
% (and its generator at make_convergence.py) so this can be reverted. The
% LAYOUT IS UNCHANGED on purpose: same axes and ranges, same six series, same
% solid-full / dashed-LoRA encoding, same filled-vs-hollow markers, same two
% legends in the same corners. Only the drawing style moved into the family.
%
% COLOUR. The original triple was #2C5CE6 / #15803D / #BE2A2A, whose green and
% red are the pairing the paper's palette exists to avoid. These are the same
% three positions -- blue, green, warm -- taken from the Okabe-Ito CVD-safe set
% and muted to the family register, and they match the first three suite hues
% of figures/benchmark.tikz. That reuse is intentional: both are nominal
% categories with no cross-figure relation, and a reader is not invited to
% map a LIBERO suite onto a policy. Neither is flcol or centcol, which stay
% reserved for the federated-vs-centralized contrast.
%
% Thin ink takes a darker step of the fill written as a mix, per the repo rule.
\definecolor{polA}{HTML}{1F6FA8}     % pi_0.5    -- blue
\definecolor{polB}{HTML}{2E8B6F}     % GR00T     -- bluish green
\definecolor{polC}{HTML}{B4653A}     % SmolVLA   -- vermillion, muted
\definecolor{shadeink}{HTML}{16191F}
\definecolor{stageink}{HTML}{444B59}

\begin{tikzpicture}[x=1cm, y=1cm,
  ln/.style={line width=0.9pt},
  lora/.style={dash pattern=on 2.4pt off 1.6pt},
  mk/.style={draw=shadeink, line width=0.4pt},
  tick/.style={font=\scriptsize, text=stageink},
  lgd/.style={font=\scriptsize, text=stageink, anchor=west},
]

% ---- grid ----
\foreach \y in {0.9104, 1.8208, 2.7312, 3.6416, 4.5520}
  \draw[gray!25, line width=0.4pt] (0,\y) -- (6.60,\y);
\foreach \x in {0.9429, 1.8857, 2.8286, 3.7714, 4.7143, 5.6571, 6.6000}
  \draw[gray!25, line width=0.4pt] (\x,0) -- (\x,4.55);

% ---- axes ----
\draw[stageink!60, line width=0.6pt] (0,0) -- (6.60,0);
\draw[stageink!60, line width=0.6pt] (0,0) -- (0,4.55);
\foreach \y/\v in {0.0000/0, 0.9104/20, 1.8208/40, 2.7312/60, 3.6416/80, 4.5520/100} {
  \draw[stageink!60, line width=0.6pt] (-0.07,\y) -- (0,\y);
  \node[tick, anchor=east] at (-0.12,\y) {\v};}
\foreach \x/\v in {0.0000/20, 0.9429/40, 1.8857/60, 2.8286/80, 3.7714/100, 4.7143/120, 5.6571/140, 6.6000/160} {
  \draw[stageink!60, line width=0.6pt] (\x,-0.07) -- (\x,0);
  \node[tick, anchor=north] at (\x,-0.13) {\v};}
\node[anchor=north, font=\footnotesize, text=stageink] at (3.30,-0.52)
  {federated rounds};
\node[rotate=90, anchor=south, font=\footnotesize, text=stageink]
  at (-0.80,2.28) {LIBERO success (\%)};

% ---- marker shapes: filled = full fine-tuning, hollow = LoRA -------------
% Distinct names: \circ and \sq would collide with existing control sequences.
% #3 is the fill, #4 the edge. A hollow LoRA marker takes its OWN series
% colour as the edge, matching the dashed line it sits on; a filled one keeps
% the neutral shadeink edge the family uses, which the fill needs for relief.
\newcommand{\cvcirc}[4]{\draw[line width=0.4pt, draw=#4, fill=#3] (#1,#2) circle (0.070);}
\newcommand{\cvsq}[4]{\draw[line width=0.4pt, draw=#4, fill=#3] ({#1-0.062},{#2-0.062}) rectangle ({#1+0.062},{#2+0.062});}
\newcommand{\cvtri}[4]{\draw[line width=0.4pt, draw=#4, fill=#3] ({#1-0.075},{#2-0.055}) -- ({#1+0.075},{#2-0.055}) -- (#1,{#2+0.085}) -- cycle;}

% ---- pi05 ----
\draw[ln, draw=polA!70!black] (0.0000,3.8578) -- (0.9429,4.1878) -- (1.8857,4.3130) -- (2.8286,4.4041) -- (3.7714,4.4382);
\draw[ln, draw=polA!70!black, lora] (0.0000,3.5961) -- (0.9429,4.1196) -- (1.8857,4.1651) -- (2.8286,4.2903) -- (3.7714,4.4382);
\cvcirc{0.0000}{3.8578}{polA}{shadeink}
\cvcirc{0.9429}{4.1878}{polA}{shadeink}
\cvcirc{1.8857}{4.3130}{polA}{shadeink}
\cvcirc{2.8286}{4.4041}{polA}{shadeink}
\cvcirc{3.7714}{4.4382}{polA}{shadeink}
\cvcirc{0.0000}{3.5961}{white}{polA!70!black}
\cvcirc{0.9429}{4.1196}{white}{polA!70!black}
\cvcirc{1.8857}{4.1651}{white}{polA!70!black}
\cvcirc{2.8286}{4.2903}{white}{polA!70!black}
\cvcirc{3.7714}{4.4382}{white}{polA!70!black}

% ---- groot ----
\draw[ln, draw=polB!70!black] (0.0000,1.2404) -- (0.9429,2.7084) -- (1.8857,3.6530) -- (2.8286,3.9033) -- (3.7714,4.1423) -- (4.7143,4.0968) -- (5.6571,4.0740);
\draw[ln, draw=polB!70!black, lora] (0.0000,1.0128) -- (0.9429,1.9574) -- (1.8857,2.9360) -- (2.8286,3.4026) -- (3.7714,3.6644) -- (4.7143,3.8009) -- (5.6571,3.9602);
\cvsq{0.0000}{1.2404}{polB}{shadeink}
\cvsq{0.9429}{2.7084}{polB}{shadeink}
\cvsq{1.8857}{3.6530}{polB}{shadeink}
\cvsq{2.8286}{3.9033}{polB}{shadeink}
\cvsq{3.7714}{4.1423}{polB}{shadeink}
\cvsq{4.7143}{4.0968}{polB}{shadeink}
\cvsq{5.6571}{4.0740}{polB}{shadeink}
\cvsq{0.0000}{1.0128}{white}{polB!70!black}
\cvsq{0.9429}{1.9574}{white}{polB!70!black}
\cvsq{1.8857}{2.9360}{white}{polB!70!black}
\cvsq{2.8286}{3.4026}{white}{polB!70!black}
\cvsq{3.7714}{3.6644}{white}{polB!70!black}
\cvsq{4.7143}{3.8009}{white}{polB!70!black}
\cvsq{5.6571}{3.9602}{white}{polB!70!black}

% ---- smolvla ----
\draw[ln, draw=polC!70!black] (0.0000,0.5690) -- (0.9429,1.4794) -- (1.8857,2.5377) -- (2.8286,2.9133) -- (3.7714,3.2774) -- (4.7143,3.3685) -- (5.6571,3.4140) -- (6.6000,3.4254);
\draw[ln, draw=polC!70!black, lora] (0.0000,0.4324) -- (0.9429,0.9332) -- (1.8857,1.3315) -- (2.8286,1.6956) -- (3.7714,1.9915) -- (4.7143,2.1622) -- (5.6571,2.4353) -- (6.6000,2.5150);
\cvtri{0.0000}{0.5690}{polC}{shadeink}
\cvtri{0.9429}{1.4794}{polC}{shadeink}
\cvtri{1.8857}{2.5377}{polC}{shadeink}
\cvtri{2.8286}{2.9133}{polC}{shadeink}
\cvtri{3.7714}{3.2774}{polC}{shadeink}
\cvtri{4.7143}{3.3685}{polC}{shadeink}
\cvtri{5.6571}{3.4140}{polC}{shadeink}
\cvtri{6.6000}{3.4254}{polC}{shadeink}
\cvtri{0.0000}{0.4324}{white}{polC!70!black}
\cvtri{0.9429}{0.9332}{white}{polC!70!black}
\cvtri{1.8857}{1.3315}{white}{polC!70!black}
\cvtri{2.8286}{1.6956}{white}{polC!70!black}
\cvtri{3.7714}{1.9915}{white}{polC!70!black}
\cvtri{4.7143}{2.1622}{white}{polC!70!black}
\cvtri{5.6571}{2.4353}{white}{polC!70!black}
\cvtri{6.6000}{2.5150}{white}{polC!70!black}

% ---- legend, below the axis ---------------------------------------------
\draw[ln, draw=polA!70!black] (0.55,-1.26) -- (1.13,-1.26);
\cvcirc{0.84}{-1.26}{polA}{shadeink}
\node[lgd] at (1.23,-1.26) {$\pi_{0.5}$};
\draw[ln, draw=polB!70!black] (1.98,-1.26) -- (2.56,-1.26);
\cvsq{2.27}{-1.26}{polB}{shadeink}
\node[lgd] at (2.66,-1.26) {GR00T~N1.7};
\draw[ln, draw=polC!70!black] (4.31,-1.26) -- (4.89,-1.26);
\cvtri{4.60}{-1.26}{polC}{shadeink}
\node[lgd] at (4.99,-1.26) {SmolVLA};
\draw[ln, draw=stageink!75] (1.38,-1.68) -- (1.96,-1.68);
\node[lgd] at (2.06,-1.68) {full fine-tuning};
\draw[ln, draw=stageink!75, lora] (3.89,-1.68) -- (4.47,-1.68);
\node[lgd] at (4.57,-1.68) {LoRA};

\end{tikzpicture}

%% file: figures/real.tikz
\definecolor{fedcol}{HTML}{5C9FD4}      % federated   -- light blue
\definecolor{pooledcol}{HTML}{EC9A55}   % centralized -- light orange
\definecolor{shadeink}{HTML}{16191F}
\definecolor{stageink}{HTML}{444B59}
\definecolor{siteAcol}{HTML}{6A4C9C}    % site triple, shared with sites/perjoint
\definecolor{siteBcol}{HTML}{2E8B8B}
\definecolor{siteCcol}{HTML}{A23B47}

\begin{tikzpicture}[x=1cm, y=1cm,
  soft/.style={fill=shadeink, fill opacity=0.42, draw=none},
  fbar/.style={fill=fedcol,    draw=shadeink, line width=0.4pt},
  pbar/.style={fill=pooledcol, draw=shadeink, line width=0.4pt},
  val/.style={font=\tiny, text=stageink, inner sep=0pt},
]

% ===== panel frames: gridlines, ticks, group labels =========================
% \p is the panel's x origin, \w its plot width.
\foreach \p/\w in {0/6.0, 8.0/7.9} {
  \foreach \v in {0, 20, 40, 60, 80, 100} {
    \draw[gray!25, line width=0.4pt] ({\p},{\v*0.042}) -- ({\p+\w},{\v*0.042});
    \draw[stageink!60, line width=0.55pt] ({\p-0.07},{\v*0.042}) -- ({\p},{\v*0.042});
    \node[anchor=east, font=\scriptsize, text=stageink]
      at ({\p-0.12},{\v*0.042}) {\v};
  }
  \draw[stageink!60, line width=0.6pt] ({\p},0) -- ({\p+\w},0);
  \draw[stageink!60, line width=0.6pt] ({\p},0) -- ({\p},4.35);
}
\node[rotate=90, anchor=south, font=\footnotesize, text=stageink]
  at (-0.95,2.1) {cube-stacking success (\%)};

% Rule before each `Mean` group: it pools the sites to its left, so it is not
% another site and must not read as one.
\draw[gray!35, line width=0.5pt] (3.95,0) -- (3.95,4.2);
\draw[gray!35, line width=0.5pt] (13.85,0) -- (13.85,4.2);

% ===== panel (a): two clients, 3/3 task split ===============================
\foreach \x/\v in {1.37/43.3, 3.27/66.7, 5.17/55.0} {
  \fill[soft] ({\x-0.29},0) rectangle ({\x+0.15},{\v*0.042});
  \draw[fbar] ({\x-0.22},0) rectangle ({\x+0.22},{\v*0.042});
  \node[val, anchor=south] at ({\x},{\v*0.042+0.05}) {\v};
}
\foreach \x/\v in {0.83/61.7, 2.73/80.0, 4.63/70.8} {
  \fill[soft] ({\x-0.29},0) rectangle ({\x+0.15},{\v*0.042});
  \draw[pbar] ({\x-0.22},0) rectangle ({\x+0.22},{\v*0.042});
  \node[val, anchor=south] at ({\x},{\v*0.042+0.05}) {\v};
}

% ===== panel (b): three clients, 2/2/2 color-pair split =====================
\foreach \x/\v in {9.37/48.3, 11.27/56.7, 13.17/25.0, 15.07/43.3} {
  \fill[soft] ({\x-0.29},0) rectangle ({\x+0.15},{\v*0.042});
  \draw[fbar] ({\x-0.22},0) rectangle ({\x+0.22},{\v*0.042});
  \node[val, anchor=south] at ({\x},{\v*0.042+0.05}) {\v};
}
\foreach \x/\v in {8.83/65.0, 10.73/86.7, 12.63/51.7, 14.53/67.8} {
  \fill[soft] ({\x-0.29},0) rectangle ({\x+0.15},{\v*0.042});
  \draw[pbar] ({\x-0.22},0) rectangle ({\x+0.22},{\v*0.042});
  \node[val, anchor=south] at ({\x},{\v*0.042+0.05}) {\v};
}

% ===== group labels =========================================================
% The site LETTER wears the site colour of Figs. 5 and 9, so a group can be
% tied back to its rig without a second legend; `Site' and `Mean' stay in
% stageink. This is a deliberate, scoped exception to the rule that figure text
% never carries a colour -- the identifier is doing the work of a mark here, and
% these hues are the site triple, not the federated/centralized pair the bars
% themselves use.
\foreach \x/\s in {1.1/{Site \textcolor{siteAcol}{A}},
                   3.0/{Site \textcolor{siteBcol}{B}}, 4.9/Mean,
                   9.1/{Site \textcolor{siteAcol}{A}},
                   11.0/{Site \textcolor{siteBcol}{B}},
                   12.9/{Site \textcolor{siteCcol}{C}}, 14.8/Mean}
  \node[anchor=base, font=\scriptsize, text=stageink] at (\x,-0.32) {\s};

% ===== panel titles =========================================================
\node[anchor=base, font=\footnotesize, text=stageink] at (3.0,4.55)
  {(a) two clients, 3\,/\,3 task split};
\node[anchor=base, font=\footnotesize, text=stageink] at (11.95,4.55)
  {(b) three clients, 2\,/\,2\,/\,2 color-pair split};

% ===== legend ===============================================================
% Upper left of panel (a), the one region both panels leave empty. Order is
% the bar order within a group, left to right.
% Both arms are the same policy; said once, under the swatches.
\node[anchor=west, font=\scriptsize, text=stageink] at (0.12,3.48) {all \pifive{}};
\foreach \y/\sty/\lab in {4.02/pbar/{centralized},
                          3.70/fbar/{federated}} {
  \fill[soft] (0.05,\y) rectangle (0.39,{\y+0.18});
  \draw[\sty] (0.12,\y) rectangle (0.46,{\y+0.18});
  \node[anchor=west, font=\scriptsize, text=stageink] at (0.58,{\y+0.09}) {\lab};
}

\end{tikzpicture}

%% file: figures/purt.tikz
\definecolor{flcol}{HTML}{5C9FD4}     % federated   -- light blue
\definecolor{centcol}{HTML}{EC9A55}   % centralized -- light orange
\definecolor{shadeink}{HTML}{16191F}
\definecolor{stageink}{HTML}{444B59}

\begin{tikzpicture}[x=1cm, y=1cm,
  cline/.style={draw=centcol!70!black, line width=0.9pt},
  fline/.style={draw=flcol!70!black,   line width=0.9pt, dash pattern=on 2.4pt off 1.6pt},
  cdot/.style={fill=centcol, draw=shadeink, line width=0.4pt},
  fdot/.style={fill=flcol,   draw=shadeink, line width=0.4pt},
  num/.style={font=\scriptsize, text=stageink, inner sep=1.1pt},
  tick/.style={font=\scriptsize, text=stageink, anchor=east, inner sep=0pt},
  band/.style={font=\footnotesize, text=stageink, anchor=west, inner sep=0pt},
]

% ---- y axis, in three pieces with breaks between ---------------------------
\draw[stageink!60, line width=0.6pt] (0,0)    -- (0,1.50);
\draw[stageink!60, line width=0.6pt] (0,1.75) -- (0,3.25);
\draw[stageink!60, line width=0.6pt] (0,3.50) -- (0,5.00);
\foreach \yb in {1.50, 3.25} {
  \draw[stageink!60, line width=0.5pt] (-0.09,\yb+0.045) -- (0.09,\yb+0.115);
  \draw[stageink!60, line width=0.5pt] (-0.09,\yb+0.135) -- (0.09,\yb+0.205);
}
\node[rotate=90, anchor=south, font=\footnotesize, text=stageink]
  at (-0.86,2.5) {per-stage rate (\%)};

% ---- x axis ----------------------------------------------------------------
\draw[stageink!60, line width=0.6pt] (0,0) -- (5.6,0);
\foreach \x/\v in {0/0, 1.12/1, 3.36/3, 5.6/5} {
  \draw[stageink!60, line width=0.6pt] (\x,-0.07) -- (\x,0);
  \node[anchor=north, font=\scriptsize, text=stageink] at (\x,-0.14) {\v};
}
\node[anchor=north, font=\footnotesize, text=stageink] at (2.8,-0.52)
  {perturbation amplitude (cm)};

% ---- per-segment grid ------------------------------------------------------
\foreach \y in {3.8529, 4.2941, 4.7353, 2.0909, 2.4318, 2.7727, 0.4412, 0.8824, 1.3235}
  \draw[gray!22, line width=0.4pt] (0,\y) -- (5.6,\y);

% ---- y ticks, local to each segment ----------------------------------------
\foreach \y/\v in {3.8529/80, 4.2941/85, 4.7353/90, 2.0909/35, 2.4318/40, 2.7727/45, 0.4412/5, 0.8824/10, 1.3235/15} {
  \draw[stageink!60, line width=0.6pt] (-0.07,\y) -- (0,\y);
  \node[tick] at (-0.12,\y) {\v};
}

% ---- contact segment ---------------------------------------------
\draw[cline] (0.00,4.6912) -- (1.12,4.3206) -- (3.36,4.3824) -- (5.60,3.9765);
\draw[fline] (0.00,4.3559) -- (1.12,4.2147) -- (3.36,4.2500) -- (5.60,3.7382);
\foreach \x/\y in {0.00/4.6912, 1.12/4.3206, 3.36/4.3824, 5.60/3.9765} \draw[cdot] (\x,\y) circle (0.062);
\foreach \x/\y in {0.00/4.3559, 1.12/4.2147, 3.36/4.2500, 5.60/3.7382} \draw[fdot] (\x,\y) circle (0.062);
\node[num, anchor=south west] at (0.09,4.7662) {89.5};
\node[num, anchor=south] at (1.12,4.3956) {85.3};
\node[num, anchor=south] at (3.36,4.4574) {86.0};
\node[num, anchor=south east] at (5.51,4.0515) {81.4};
\node[num, anchor=north west] at (0.09,4.2809) {85.7};
\node[num, anchor=north] at (1.12,4.1397) {84.1};
\node[num, anchor=north] at (3.36,4.1750) {84.5};
\node[num, anchor=north east] at (5.51,3.6632) {78.7};

% ---- transport segment -------------------------------------------
\draw[cline] (0.00,2.9295) -- (1.12,2.6636) -- (3.36,2.6909) -- (5.60,2.2409);
\draw[fline] (0.00,2.2409) -- (1.12,1.9273) -- (3.36,2.1386) -- (5.60,2.0023);
\foreach \x/\y in {0.00/2.9295, 1.12/2.6636, 3.36/2.6909, 5.60/2.2409} \draw[cdot] (\x,\y) circle (0.062);
\foreach \x/\y in {0.00/2.2409, 1.12/1.9273, 3.36/2.1386, 5.60/2.0023} \draw[fdot] (\x,\y) circle (0.062);
\node[num, anchor=south west] at (0.09,3.0045) {47.3};
\node[num, anchor=south] at (1.12,2.7386) {43.4};
\node[num, anchor=south] at (3.36,2.7659) {43.8};
\node[num, anchor=south east] at (5.51,2.3159) {37.2};
\node[num, anchor=north west] at (0.09,2.1659) {37.2};
\node[num, anchor=north] at (1.12,1.8523) {32.6};
\node[num, anchor=north] at (3.36,2.0636) {35.7};
\node[num, anchor=north east] at (5.51,1.9273) {33.7};

% ---- success segment ---------------------------------------------
\draw[cline] (0.00,1.2000) -- (1.12,1.0235) -- (3.36,0.7853) -- (5.60,0.7853);
\draw[fline] (0.00,0.6882) -- (1.12,0.7500) -- (3.36,0.3441) -- (5.60,0.5118);
\foreach \x/\y in {0.00/1.2000, 1.12/1.0235, 3.36/0.7853, 5.60/0.7853} \draw[cdot] (\x,\y) circle (0.062);
\foreach \x/\y in {0.00/0.6882, 1.12/0.7500, 3.36/0.3441, 5.60/0.5118} \draw[fdot] (\x,\y) circle (0.062);
\node[num, anchor=south west] at (0.09,1.2750) {13.6};
\node[num, anchor=south] at (1.12,1.0985) {11.6};
\node[num, anchor=south] at (3.36,0.8603) {8.9};
\node[num, anchor=south east] at (5.51,0.8603) {8.9};
\node[num, anchor=north west] at (0.09,0.6132) {7.8};
\node[num, anchor=north] at (1.12,0.6750) {8.5};
\node[num, anchor=north] at (3.36,0.2691) {3.9};
\node[num, anchor=north east] at (5.51,0.4368) {5.8};
% ---- band names ------------------------------------------------------------
\node[band] at (5.76,4.25) {contact};
\node[band] at (5.76,2.50) {transport};
\node[band] at (5.76,0.75) {success};

% ---- legend, above the plot -------------------------------------------------
% Clears the tallest ink below it: the 89.5 label tops out at about 4.94.
\draw[cline] (1.02,5.28) -- (1.50,5.28);
\draw[cdot]  (1.26,5.28) circle (0.062);
\node[anchor=west, font=\scriptsize, text=stageink] at (1.60,5.28) {centralized};
\draw[fline] (3.20,5.28) -- (3.68,5.28);
\draw[fdot]  (3.44,5.28) circle (0.062);
\node[anchor=west, font=\scriptsize, text=stageink] at (3.78,5.28) {federated};

\end{tikzpicture}

%% file: sections/conclusion.tex
\section{Conclusion}\label{sec:conclusion}
This study shows that federated fine-tuning is a promising,
privacy-preserving route for Vision-Language-Action models to keep
improving after release: when the policy's pretraining is strong,
it matches or approaches centralized fine-tuning without any
demonstration leaving its site, so homes, labs, and factories could
in principle train a shared policy.

Realizing this prospect requires resolving the open problems the
study measured. LoRA reduces the transmission payload only by paying
rounds, so communication-efficient federation converges slowly.
None of the evaluated aggregation algorithms improves on FedAvg, and
weakly pretrained policies still trail their centralized
counterparts. On physical robots, cross-site heterogeneity is
stronger than simulated benchmarks express, especially in the action
space. We leave these problems to the community and release
\decentvla{}, with the configuration behind every reported result,
as a reproducible benchmark on which to pursue them.

%% file: sections/acknowledgment.tex
\section*{Acknowledgment}
This work was sponsored by NVIDIA and Cisco. Experiments were run on
Isambard-AI, part of the UK's AI Research Resource, funded by UK
Research and Innovation [ST/AIRR/I-A-I/1023].

%% file: sections/references.tex
% Manual thebibliography (no BibTeX), matching the repo convention.
%
% ORDERED BY FIRST CITATION IN TEXT -- IEEE requires this; do NOT re-sort by
% topic. When adding a reference, insert it at its first-citation position.
% Re-check ordering with:
%   grep -oh '\\cite[^}]*}' sections/*.tex   (in \input order from main.tex)
%
% Metadata re-verified 2026-09-12 against arXiv, PMLR (CoRL/ICML/AISTATS),
% OpenReview/ICLR, ACL Anthology, ACM DL, IEEE/CrossRef DOI records, official
% RSS/ICCV program pages, and vendor pages; every entry checked field-by-field.
% Corrected in that pass: fedit missing co-author (Y. Zhou); flame upgraded
% arXiv->IROS 2025; pi05 upgraded to its CoRL 2025 pages; full list re-sorted
% to first-citation order after the 07-27 restructure. arXiv/model-card
% citations kept ONLY where no peer-reviewed venue exists as of 2026-09-12
% (SmolVLA, GR00T N1, GR00T N1.7, Flower, FedPer, LIBERO-PRO, pi_0.7,
% RoboLab); re-check LIBERO-PRO and SmolVLA at camera-ready. Still missing
% throughout: DOIs.
%%%%%%%%%%%%%%%%%%%%%%%%%%%%%%%%%%%%%%%%%%%%%%%%%%%%%%%%%%%%%%%%%%%%%%%%%%%%%%%%